\documentclass[10pt,journal,compsoc]{IEEEtran}

\ifCLASSOPTIONcompsoc
  \usepackage[nocompress]{cite}
\else
  \usepackage{cite}
\fi

\ifCLASSINFOpdf
\else
\fi

\usepackage{times}
\usepackage{epsfig}
\usepackage{graphicx}
\usepackage{amsmath}
\usepackage{amssymb}
\usepackage{booktabs}
\graphicspath{{cambriafig/}}
\usepackage[pagebackref=true,breaklinks=true,colorlinks,bookmarks=false]{hyperref}
\usepackage{float}
\usepackage{multirow}
\usepackage{tabularx}
\newcommand{\bs}{\boldsymbol}

\begin{document}
%
\title{PMP-Net++: Point Cloud Completion by Transformer-Enhanced \\Multi-step Point Moving Paths}
%
%
%
%
\author{Xin~Wen, Peng~Xiang, Zhizhong~Han, Yan-Pei~Cao, Pengfei~Wan, Wen~Zheng, Yu-Shen~Liu~\IEEEmembership{Member,~IEEE,}

\IEEEcompsocitemizethanks{
\IEEEcompsocthanksitem Xin Wen and Peng Xiang are with the School of Software, Tsinghua University, Beijing, China. Xin Wen is also with JD Logistics, JD.com, China. E-mail: wenxin16@jd.com, xp20@mails.tsinghua.edu.cn
\IEEEcompsocthanksitem Zhizhong Han is with the Department of Computer Science, Wayne State University, USA. E-mail: h312h@wayne.edu
\IEEEcompsocthanksitem Yu-Shen Liu is with the School of Software, BNRist, Tsinghua University, Beijing, China. E-mail: liuyushen@tsinghua.edu.cn
\IEEEcompsocthanksitem Yan-Pei Cao, Pengfei Wan and Wen Zheng are with the Y-tech, Kuaishou Technology, Beijing, China. Email: caoyanpei@gmail.com, \{wanpengfei, zhengwen\}@kuaishou.com}
\thanks{Yu-Shen Liu is the corresponding author. This work was supported by National Key R\&D Program of China (2020YFF0304100), the National Natural Science Foundation of China (62072268), and in part by Tsinghua-Kuaishou Institute of Future Media Data. Code is available at \url{https://github.com/diviswen/PMP-Net}.}
}

%
%

\markboth{Journal of \LaTeX\ Class Files,~Vol.~14, No.~8, August~2015}%
{Shell \MakeLowercase{\textit{et al.}}: Bare Demo of IEEEtran.cls for Computer Society Journals}
%



\IEEEtitleabstractindextext{%
\begin{abstract}
Point cloud completion concerns to predict missing part for incomplete 3D shapes. A common strategy is to generate complete shape according to incomplete input. However, unordered nature of point clouds will degrade generation of high-quality 3D shapes, as detailed topology and structure of unordered points are hard to be captured during the generative process using an extracted latent code. We address this problem by formulating completion as point cloud deformation process. Specifically, we design a novel neural network, named PMP-Net++, to mimic behavior of an earth mover. It moves each point of incomplete input to obtain a complete point cloud, where total distance of point moving paths (PMPs) should be the shortest. Therefore, PMP-Net++ predicts unique PMP for each point according to constraint of point moving distances. The network learns a strict and unique correspondence on point-level, and thus improves quality of predicted complete shape. Moreover, since moving points heavily relies on per-point features learned by network, we further introduce a transformer-enhanced representation learning network, which significantly improves completion performance of PMP-Net++. We conduct comprehensive experiments in shape completion, and further explore application on point cloud up-sampling, which demonstrate non-trivial improvement of PMP-Net++ over state-of-the-art point cloud completion/up-sampling methods.
\end{abstract}

}

\maketitle

\IEEEdisplaynontitleabstractindextext

%
\IEEEpeerreviewmaketitle

\IEEEraisesectionheading{\section{Introduction}\label{sec:introduction}}

%
%
%
%
\IEEEPARstart{A}{s} one of the widely used 3D shape representations, point clouds can be easily obtained through depth cameras or other 3D scanning devices. Due to the limitations of view-angles or occlusions of 3D scanning devices, the raw point clouds are usually sparse and incomplete \cite{wen2020sa}. Therefore, a shape completion/consolidation process is usually required to generate the missing regions of 3D shape for the downstream 3D computer vision applications like classification \cite{han20193dviewgraph,han2019parts,wen2020point2spatialcapsule,han2019seqviews2seqlabels,liu2019sequence,liu2020lrc}, segmentation \cite{Wen2020MM,jiang2020pointgroup} and other visual analysis \cite{yuan2021survey}.

\begin{figure*}[!t]
  \centering
  \includegraphics[width=\textwidth]{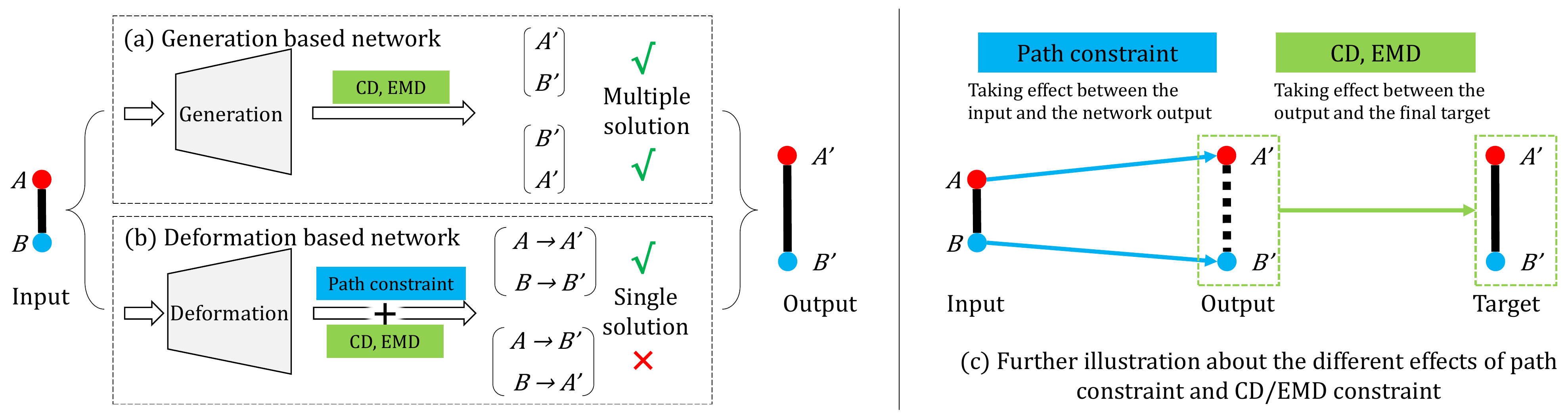}
  \caption{Illustration of the differences between the generation based methods and the deformation based methods, where the task is to complete a short line $AB$ to a long line $A'B'$ (in (a) and (b)). The differences of the effect between the path constraint and the widely used CD/EMD is further illustrated in (c).
  }
  \label{fig:problem}
\end{figure*}

In this paper, we focus on the completion task for 3D objects represented by point clouds, where the missing parts are caused by self-occlusion due to the view angle of scanner. Most of the previous methods formulate the point cloud completion as a point cloud generation problem \cite{dai2017shape,wen2020sa,yuan2018pcn,tchapmi2019topnet}, where an encoder-decoder framework is usually adopted to extract a latent code from the input incomplete point cloud, and decode the extracted latent code into a complete point cloud. Benefiting from the deep neural network based point cloud learning methods, the point cloud completion methods along this line have made huge progress in the last few years \cite{wen2020sa,tchapmi2019topnet}. However, the generation of point clouds remains a difficult task using deep neural network, because the unordered nature of point clouds makes the generative model difficult to capture the detailed topology or structure among discrete points \cite{tchapmi2019topnet}. Therefore, the performance of generative models based point clouds completion is still unsatisfactory.

To improve the point cloud completion performance, in this paper, we propose a novel deep neural network, named PMP-Net++, to formulate the task of point cloud completion from a new perspective. Different from the generative model that directly predicts the coordinations of all points in 3D space, the PMP-Net++ learns to move the points from the source 3D shape to the target one.
Through the point moving process, the PMP-Net++ learns the point-level correspondences between the source point cloud and the target, which captures the detailed topology and structure relationships between the two point clouds. On the other hand, there are many possible solutions to move points from the source to the target, which will make the network difficult to train well.
Therefore, in order to encourage the network to learn a unique optimal arrangement of point moving path, we take the inspiration from the Earth Mover's Distance (EMD) and propose to regularize a transformer-enhanced Point-Moving-Path Network (named PMP-Net++) under the constraint of the total point moving distances (PMDs), which guarantees the uniqueness of path arrangement between the source point cloud and the target one.

\begin{figure}[!t]
  \centering
  \includegraphics[width=\columnwidth]{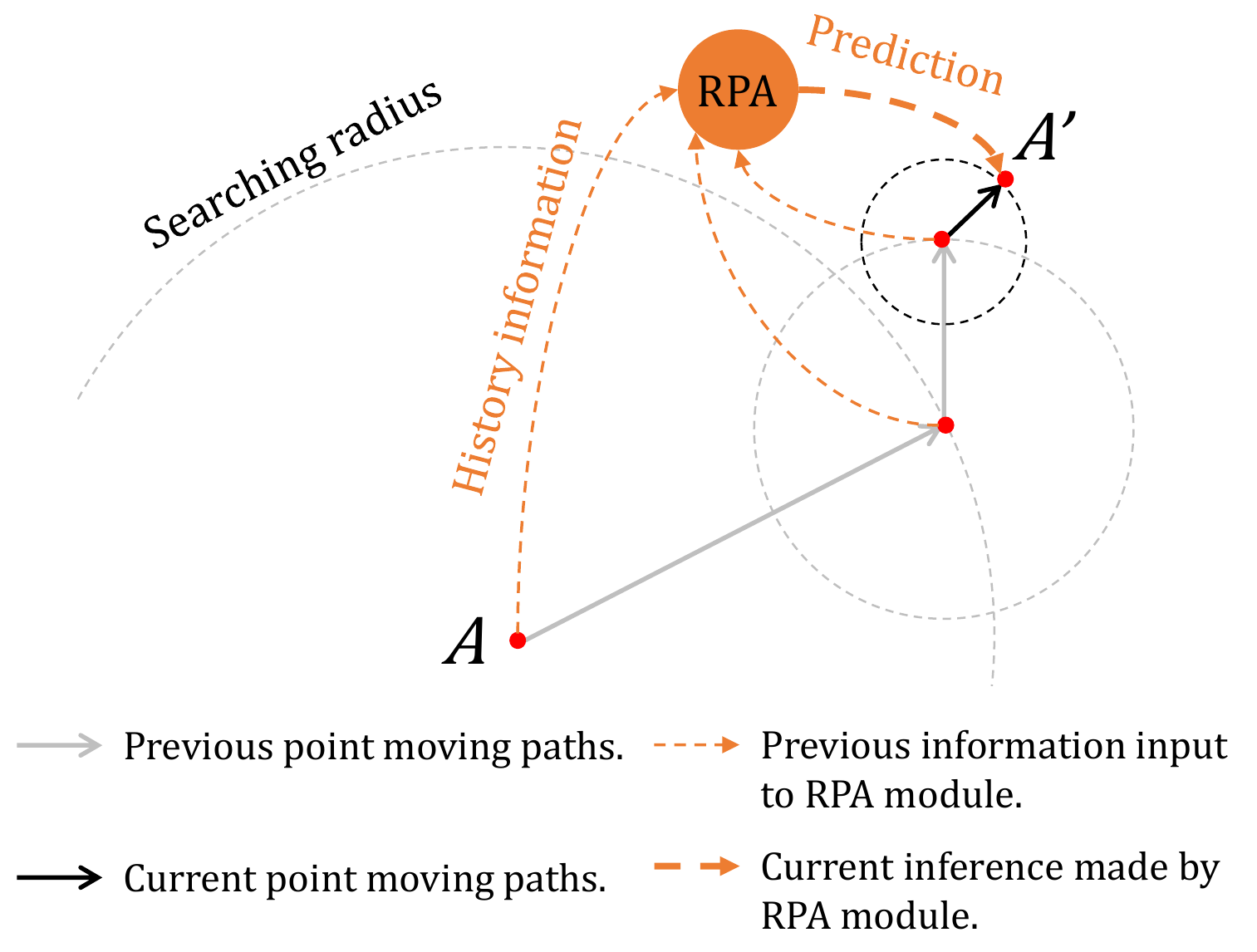}
  \caption{Illustration of path searching with multiple steps under the coarse-to-fine searching radius. The PMP-Net++ moves point $A$ to point $A'$ by three steps, with each step reducing its searching radius, and looking back to consider the moving history in order to decide the next place to move.
  }
  \label{fig:searching_illustration}
\end{figure}

A detailed illustration is given in Figure \ref{fig:problem}. Taking the task of completing short line $AB$ to the long line $A'B'$ for example, the generation based neural network aims to predict the coordinates of $A'B'$, which is usually optimized by the chamfer distance (CD) or earth mover's distance (EMD). However, due to the discrete nature of point cloud data, the target matrix representing the line $A'B'$ has multiple arrangements, all of which meet the minimum loss of CD and EMD constraints. As a result, there are multiple optimal targets for the network, which cannot guide the network to learn the detailed shape correspondence between the short line $AB$ and the long line $A'B'$. In contrast, shape deformation based neural network can potentially establish a direct and point-wise correspondence between the source input ($AB$) and the target output ($A'B'$), with the guidance of path constraint. The reason is further illustrated in Figure \ref{fig:problem} (c). The path constraint takes effect between the input and the output of the network. It regularizes the point moving path and optimizes the network to predict a unique displacement for each point. Under such circumstance, the output of network is the displacement, which is locally related to each start point and end point. On the other hand, the output of generation based network is the coordinate matrix representing line $A'B'$, and its only source of supervision is the overall shape constraints CD/EMD, which takes effect between the output and the target ground truth.

Moreover, in order to predict the point moving path more accurately, we propose a multi-step path searching strategy to continuously refine the point moving path under multi-scaled searching radius. Specifically, as shown in Figure \ref{fig:searching_illustration}, the path searching is repeated for multiple steps in a coarse-to-fine manner. Each step will take the previously predicted paths into consideration, and then plan its next move path according to the previous paths. To record and aggregate the history information of point moving path, we take the inspiration from Gated Recurrent Unit (GRU) to propose a novel Recurrent Path Aggregation (RPA) module. It can memorize and aggregate the route sequence for each point, and combine the previous information with the current location of point to predict the direction and the length for the next move. By reducing the searching radius step-by-step, PMP-Net++ can consistently refine a more and more accurate path for each point to move from its original position on the incomplete point cloud to the target position on the complete point cloud.

The PMP-Net++ proposed in this paper is an enhanced extension of our latest work PMP-Net~\cite{wen2021pmp}. We find that the point features learned in the moving procedure plays the key role during the prediction of high quality complete shape. And the PointNet++\cite{qi2017pointnet++} based backbone used in PMP-Net cannot provide more discriminative point features, due to its max-pooling based feature aggregation strategy \cite{wen2020point2spatialcapsule}. Therefore, inspired by the recent success of Transformer in point cloud representation learning \cite{zhao2020point}, we introduce a novel transformer based framework into PMP-Net++ to enhance the point features learned by our network, which aims to predict a more accurate displacement for each point.
In all, the main contributions of our work can be summarized as follows.
\begin{itemize}
  \item We propose a novel network for point cloud completion task, named PMP-Net++, to move each point on the incomplete shape to the complete one to achieve a highly accurate point cloud completion. Compared with previous generative completion methods, PMP-Net++ has the ability to learn more detailed topology and structure relationships between incomplete shapes and complete ones, by learning the point-level correspondence through point moving path prediction.
  \item We propose to learn a unique point moving path arrangement between incomplete and complete point clouds, by regularizing the network using the constraint of Earth Mover's Distance. As a result, the network will not be confused by multiple solutions of moving points, and finally predicts a meaningful point-wise correspondence between the source and target point clouds.
  \item We propose to search point moving path with multiple steps in a coarse-to-fine manner. Each step will decide the next move based on the aggregated information from the previous paths and its current location, using the proposed Recurrent Path Aggregation (RPA) module.
  \item Compared with our latest PMP-Net, we further introduce a transformer-enhanced point cloud network into PMP-Net++ to improve the point feature learning. We conduct comprehensive experiments on Completion3D\cite{tchapmi2019topnet} and PCN\cite{yuan2018pcn} datasets, and further explore the application on point cloud up-sampling, all of which demonstrate the non-trivial improvement of PMP-Net++ over the state-of-the-art point cloud completion/up-sampling methods (including our latest PMP-Net).
\end{itemize}

\section{Related Work}
\label{sec:rel}

The deep learning technology in 3D reconstruction \cite{Han2020TIP,Han2020Sdrwr,Han2020ECCV,Jiang2019SDFDiffDRcvpr,Han2020Sdrwr} and representation learning \cite{9318534,han2019multi,liu2019l2g,han20193d2seqviews} have boosted the research of 3D shape completion, which can be roughly divided into two categories. (1) Traditional 3D shape completion methods \cite{sung2015data,berger2014state,thanh2016field,wei2019local} usually formulate hand-crafted features such as surface smoothness or symmetry axes to infer the missing regions, while some other methods \cite{shao2012interactive,kalogerakis2012probabilistic,martinovic2013bayesian,shen2012structure} consider the aid of large-scale complete 3D shape datasets, and perform searching to find the similar patches to fill the incomplete regions of 3D shapes. (2) Deep learning based methods \cite{zhang2020detail,sarmad2019rl,huang2020pf,hutaoaaai2020}, on the other hand, exploit the powerful representation learning ability to extract geometric features from the incomplete input shapes, and directly infer the complete shape according to the extracted features. Those learnable methods do not require the predefined hand-crafted features in contrast with traditional completion methods, and can better utilize the abundant shape information lying in the large-scale completion datasets. The proposed PMP-Net++ also belongs to the deep learning based method, where the methods along this line can be further categorized and detailed as below.

\subsection{Volumetric aided shape completion}
The representation learning ability of convolutional neural network (CNN) has been widely used in 2D computer vision research, and the studies concerning application of 2D image inpainting have been continuously surging in recent years. A intuitive idea for 3D shape completion can be directly borrowed from the success of 2D CNN in image inpainting research \cite{yu2018generative,yeh2017semantic,liu2018image}, extending it into 3D space. Recently, several volumetric aided shape completion methods, which are based on 3D CNN structure, have been developed. Note that we use the term \emph{``volumetric aided''} to describe this kind of methods, because the 3D voxel is usually not the final output of the network. Instead, the predicted voxel will be further refined and converted into other representations like mesh \cite{dai2017shape} or point cloud \cite{xie2020grnet}, in order to produce more detailed 3D shapes. Therefore, the voxel is more like an intermediate aid to help the completion network infer the complete shape. Notable works along this line like 3D-EPN \cite{dai2017shape} and GRNet \cite{xie2020grnet} have been proposed to reconstruct the complete 3D voxel in a coarse-to-fine manner. They first predict a coarse complete shape using 3D CNN under an encoder-decoder framework, and then refine the output using similar patches selected from a complete shape dataset \cite{dai2017shape} or by further reconstructing the detailed point cloud according to the output voxel \cite{xie2020grnet}. Also, there are some studies that consider purely volumetric data for shape completion task. For example, Han et. al \cite{han2017high} proposed to directly generate the high-resolution 3D volumetric shape, by simultaneously inferring global structure and local geometries to predict the detailed complete shape. Stutz et. al \cite{stutz2018learning} proposed a variational auto-encoder based method to complete the 3D voxel under weak supervision. Despite the fascinating ability of 3D CNN for feature learning, the computational cost which is cubic to the resolution of input voxel data makes it difficult to process fine-grained shapes \cite{wen2020sa}.

\subsection{Point cloud based shape completion}
There is a growing attention on the task of point cloud based shape completion \cite{xin2021c4c,tchapmi2019topnet,wen2020sa,hu2019render4completion} in recent years. Since point cloud is a direct output form of many 3D scanning devices, and the storage and process of point clouds require much less computational cost than volumetric data, many recent studies consider to perform direct completion on 3D point clouds. Enlighten from the improvement of point cloud representation learning \cite{qi2017pointnet++,qi2017pointnet}, previous methods like TopNet \cite{tchapmi2019topnet}, PCN \cite{yuan2018pcn} and SA-Net \cite{wen2020sa} formulate the solution as a generative model under an encoder-decoder framework. They adopted encoder like PointNet \cite{qi2017pointnet} or PointNet++ \cite{qi2017pointnet++} to extract the global feature from the incomplete point cloud, and use a decoder to infer the complete point cloud according to the extracted features. Compare to PCN \cite{yuan2018pcn}, TopNet \cite{tchapmi2019topnet} improved the structure of decoder in order to implicitly model and generate point cloud in a rooted tree architecture \cite{tchapmi2019topnet}. SA-Net \cite{wen2020sa} took one step further to preserve and convey the detailed geometric information of incomplete shape into the generation of complete shape through skip-attention mechanism. Other notable methods like RL-GAN-Net \cite{sarmad2019rl}, Render4Completion \cite{hu2019render4completion} and VRCNet \cite{pan2021vrc} focused on the framework of adversarial learning and variational auto-encoder to improve the reality and consistency of the generated complete shapes. More recently, the progressive refinement of 3D shape has become a popular idea in the research of point cloud completion (e.g. CRN \cite{wang2020cascaded}, PF-Net \cite{huang2020pf}), as it can help the network generate 3D shapes with detailed structures.

In all, most of the above methods are generative solution for point cloud completion task, and inevitably suffer from the unordered nature of point clouds, which makes it difficult to reconstruct the detailed typology or structure using a generative decoder. Therefore, in order to avoid the problem of predicting unordered data, PMP-Net++ uses a different way to reconstruct the complete point cloud, which learns to move all points from the initial input instead of directly generating the final point cloud from a latent code.

The idea of PMP-Net++ is also related to the research of 3D shape deformation \cite{yin2018p2p}, which mainly considered one-step deformation. However, the deformation between the incomplete and complete shapes is more challenging, which requires the inference of totally unknown geometries in missing regions without any other prior information. In contrast, we propose multi-step searching to encourage PMP-Net++ to infer more detailed geometric information for missing region, along with point moving distance regularization to guarantee the efficiency of multi-step inference.
\begin{figure*}[!t]
  \centering
  \includegraphics[width=\textwidth]{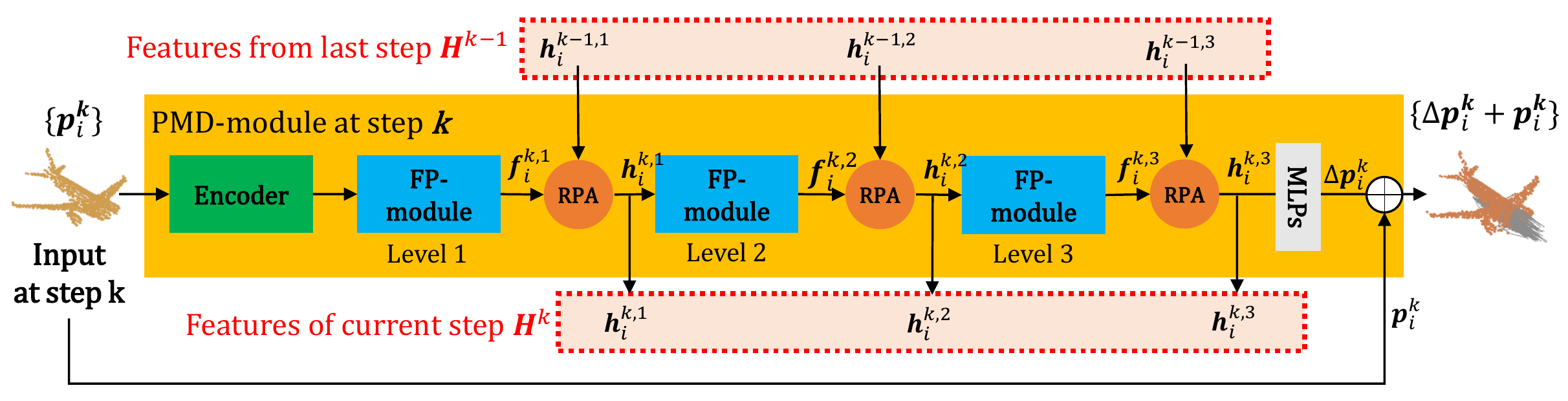}
  \caption{Detailed structure of PMD-module at step $k$. It mainly consists of three parts: (1) point cloud encoder and (2) feature prorogation module (FP-module) to extract per-point features; (3) RPA module to recurrently learn and forget the path searching information from the previous steps.
  }
  \label{fig:pmpmodule}
\end{figure*}

\begin{figure}[!t]
  \centering
  \includegraphics[width=\columnwidth]{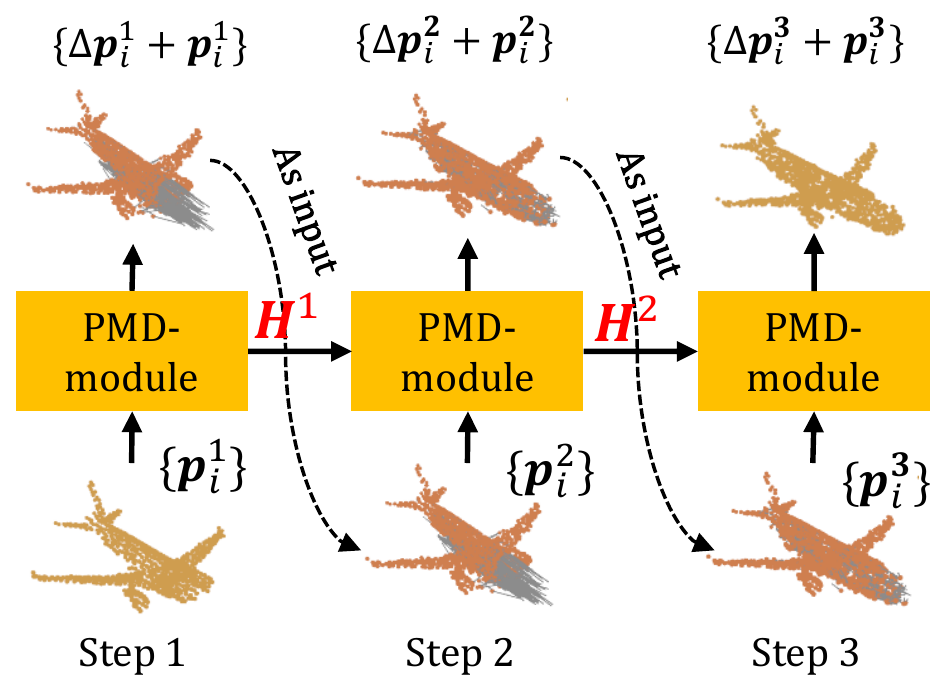}
  \caption{Illustration of path searching with multiple steps under the coarse-to-fine searching radius. The PMP-Net++ moves point A to point A' by three steps, with each step reducing its searching radius, and looking back to consider the moving history in order to decide the next place to move.
  }
  \label{fig:overview}
\end{figure}

\section{Architecture of PMP-Net}


An overview of the proposed PMP-Net++ is shown in Figure \ref{fig:overview}. The network basically consists of three parts: (1) the encoder to extract point cloud features; (2) the feature propagation module (FP-Module) to predict the point moving path for each point; (c) the RPA module recurrently fuses and integrates the current step's point features with the previous steps' path information. The details of each part will be described as below.

\subsection{Point Displacement Prediction}
\subsubsection{Multi-step framework}
An overview of the proposed PMP-Net++ is shown in Figure \ref{fig:overview}. Given an input point cloud $P=\{\bs{p}_i\}$ and a target point cloud $P'=\{\bs{p'}_j\}$. The objective of PMP-Net++ is to predict a displacement vector set $\Delta P=\{\Delta \bs{p}_i\}$, which can move each point from $P$ into the position of $P'$ such that $\{(\bs{p}_i+\Delta \bs{p}_i)\}=\{\bs{p'}_j\}$.
PMP-Net++ moves each point $\bs{p}_i$ for $K=3$ steps in total. The displacement vector for step $k$ is denoted by $\Delta\bs{p}^k_i$, so $\Delta \bs{p}_i=\sum_{k=1}^{3}\Delta\bs{p}^k_i$. For step $k$, the network takes the deformed point cloud $\{\bs{p}^{k-1}_i\}=\{\bs{p}_i+\sum_{j=1}^{k-1}\Delta\bs{p}^j_i\}$ from the last step $k-1$ as input, and calculates the new displacement vector according to the input point cloud. Therefore, the predicted shape will be consistently refined step-by-step, which finally produces a complete shape with high quality.

\begin{figure}[!t]
  \centering
  \includegraphics[width=\columnwidth]{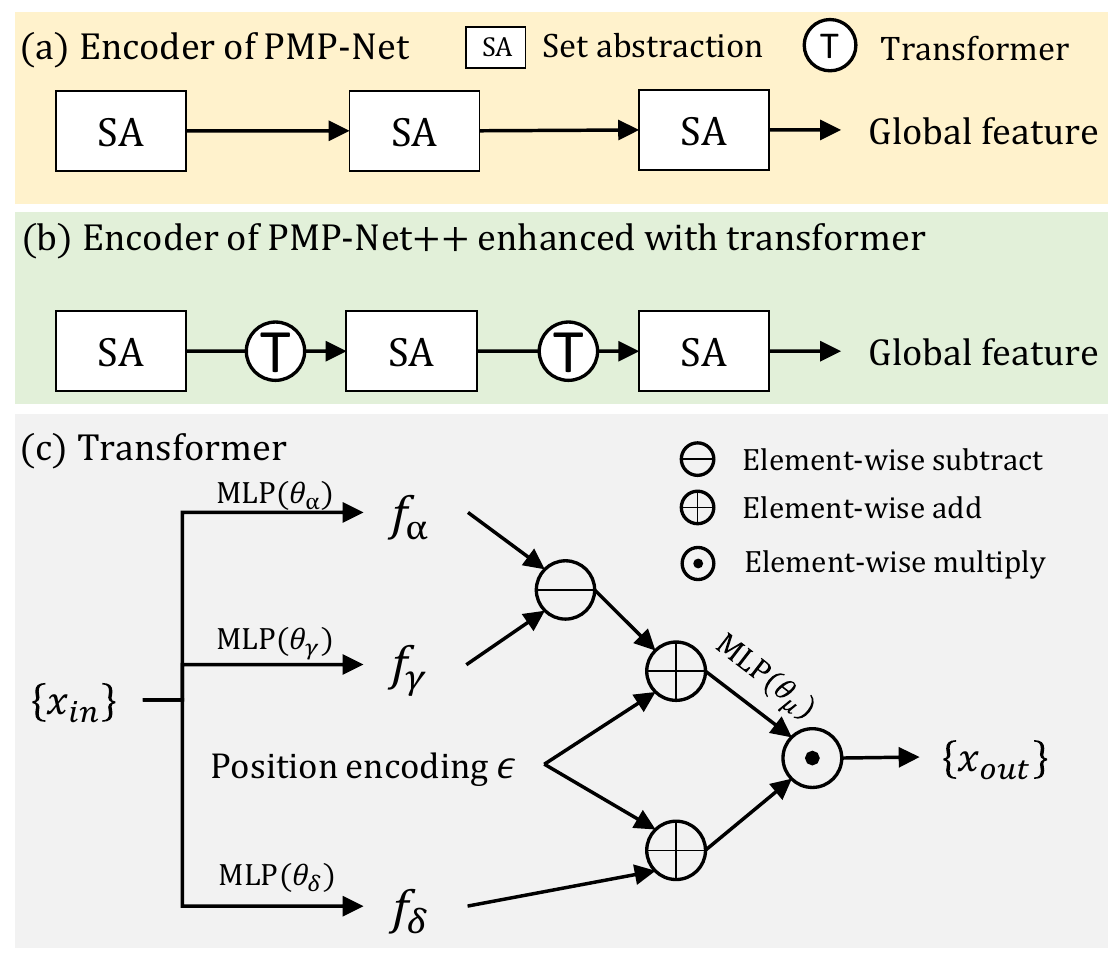}
  \caption{The architecture of encoder used in PMD-module. Moreover, we also show the comparison with previous work and the detailed structure of transformer.
  }
  \label{fig:encoder}
\end{figure}

\subsubsection{Transformer-enhanced displacement prediction}
At step $k$, in order to predict the displacement vector $\Delta\bs{p}^k_i$ for each point, we first extract per-point features from the point cloud. In the previous implementation of our PMP-Net \cite{wen2021pmp}, this is achieved by first adopting the basic framework of PointNet++ \cite{qi2017pointnet++} to extract the global feature of input the 3D shape, and then using the feature propagation module to propagate the global feature to each point in the 3D space, and finally producing per-point feature $\bs{h}^{k,l}_i$ for point $\bs{p}^k_i$. In PMP-Net++, we adopt the recent implementation success of transformer \cite{vaswani2017attention} to enhance the point feature learned by the PointNet++, where we follow the practice of Point Transformer \cite{zhao2020point}, and add an additional transformer module between each set abstraction (SA) layer of PointNet++ based encoder. The detailed structure and comparison between PMP-Net and PMP-Net++ is shown in Figure \ref{fig:encoder}. Specifically, in PMP-Net++, the local features (denoted as $\{x_{in}\}$ for convenience) learned by the previous layer of set abstraction is input to the next transformer module. In transformer module, $\{x_{in}\}$ serves as both key and query for the calculation of self-attention, according to which a set of new local features $\{x_{out}\}$ are produced through several MLPs and element-wise operations. Note that the position encoding $\epsilon$ in Figure \ref{fig:encoder} (c) is used to guide the network to learn the spatial relationships between different local features. We follow the same practice of previous work \cite{zhao2020point} to adopt the learnable position encoding, which depends on the 3D coordinates between two points $p_i$ and $p_j$:
\begin{equation}
  \epsilon = {\rm MLP}(p_i-p_j|\theta_\epsilon)
\end{equation}
Since our experimental implementation applies three levels of feature propagation to hierarchically produce per-point features (see Figure \ref{fig:pmpmodule}), we use superscript $k$ to denote the step and the subscript $l$ to denote the level in $\bs{h}^{k,l}_i$.
The per-point feature $\bs{h}^{k,l}_i$ is then concatenated with a random noise vector $\hat{\bs{x}}$, which according to \cite{yin2018p2p} can give point tiny disturbances and force it to leave its original place. Then, the final point feature $\bs{h}^{k,3}_i$ at step k and level 3 is fed into a multi-layer perceptron (MLP) followed by a hyper-tangent activation (tanh), to produce a 3-dimensional vector as the displacement vector $\Delta\bs{p}^k_i$ for point $\bs{p}^k_i$ as
\begin{equation}\label{eq:displacement}
  \Delta\bs{p}^k_i = \mathrm{tanh}(\mathrm{MLP}([\bs{h}^{k,3}_i:\hat{\bs{x}}])),
\end{equation}
where ``:'' denotes the concatenation operation.

\subsubsection{Recurrent information flow between steps}
The information of previous moves is crucial for network to decide the current move, because the previous paths can be used to infer the location of the final destination for a single point. Moreover, such information can guide the network to find the direction and distance of next move, and prevent it from changing destination during multiple steps of point moving path searching. In order to achieve this target, we propose to use a special RPA unit between each step and each level of feature propagation module, which is used to memorize the information of previous path and to infer the next position of each point. As shown in Figure \ref{fig:pmpmodule}, the RPA module in (step $k$, level $l$) takes the output $\bs{f}^{k,l-1}_i$ from the last level i-1 as input, and combines it with the feature $\bs{h}^{k-1,l}_i$ from the previous step $k-1$ at the same level $l$ to produce the feature of current level $\bs{h}^{k,l}_i$, denoted as
\begin{equation}
  \bs{h}^{k,l}_i = \mathrm{RPA}(\bs{f}^{k,l-1}_i, \bs{h}^{k-1,l}_i).
\end{equation}
The detailed structure of RPA module is described below.

\subsection{Recurrent Path Aggregation}
\begin{figure}[!t]
  \centering
  \includegraphics[width=\columnwidth]{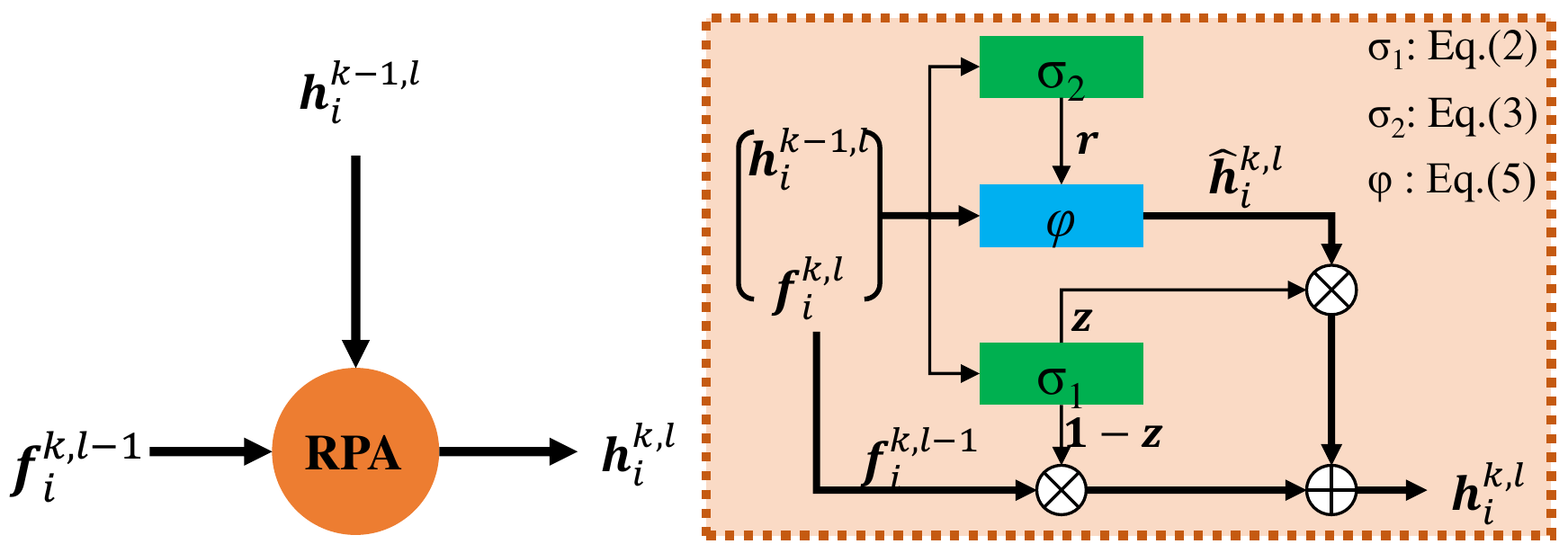}
  \caption{Detailed structure of RPA module at step $k$, level $l$.
  }
  \label{fig:rpa_structure}
\end{figure}

The detailed structure of recurrent path aggregation module is shown in Figure \ref{fig:rpa_structure}.
The previous paths of point moving can be regarded as the sequential data, where the information of each move should be selectively memorized or forgotten during the process.
Following this idea, we take the inspiration from the recurrent neural network, where we mimic the behavior of gated recurrent unit (GRU) to calculate an update gate $\bs{z}$ and reset gate $\bs{r}$ to encode and forget information, which is according to the point feature $\bs{h}^{k-1,l}_{i}$ from the last step $k-1$ and the point feature $\bs{f}^{k,l-1}_{i}$ of current step $k$. The calculation of two gates can be formulated as
\begin{equation}
  \bs{z} = \sigma(W_z[\bs{f}^{k,l-1}_{i}:\bs{h}^{k-1,l}_{i}]+\bs{b}_z),
\end{equation}
\begin{equation}
  \bs{r} = \sigma(W_r[\bs{f}^{k,l-1}_{i}:\bs{h}^{k-1,l}_{i}]+\bs{b}_r),
\end{equation}
where ${W_z, W_r}$ are weight matrix and ${\bs{b}_z,\bs{b}_z}$ are biases. $\sigma$ is the \emph{sigmoid} activation function, which predicts a value between 0 and 1 to indicate the ratio of information that allowed to pass the gate. ``:'' denotes the concatenation of two features.

Different from the standard GRU, which emphasizes more importance on the preservation of previous information when calculating the output feature $\bs{h}^{k,l}_{i}$ at current step, in RPA, we address more importance on the preservation of current input information, and propose to calculate the output feature $\bs{h}^{k,l}_{i}$ as
\begin{equation}\label{eq:hi}
  \bs{h}^{k,l}_{i} = \bs{z}\odot\bs{\hat{h}}^{k,l}_{i}+(1-\bs{z})\odot \bs{f}^{k,l-1}_{i},
\end{equation}
where $\bs{\hat{h}}^{k,l}_{i}$ is the intermediate feature of current step. It contains the preserved information from the past, which is calculated according to the current input feature. The formulation of $\bs{\hat{h}}^{k,l}_{i}$ is given as
\begin{equation}
  \bs{\hat{h}}^{k,l}_{i} = \varphi(W_h[\bs{r}\odot\bs{h}^{k-1,l}_{i}: \bs{f}^{k,l-1}_{i}]+\bs{b}_h),
\end{equation}
where $\varphi$ is \emph{relu} activation in our implementation.

The reason of fusing $\bs{\hat{h}}^{k,l}_{i}$ with $\bs{f}^{k,l-1}_{i}$ instead of $\bs{h}^{k-1,l}_{i}$ is that, compared with standard unit in RNN unit, the current location of point should have greater influence to the decision of next move. Especially, when RPA module needs to ignore the previous information which is not important in the current decision making, Eq.(\ref{eq:hi}) can easily allow RPA model to forget all history by simply pressing the update gate $\bs{z}$ to a zero-vector, and thus enables the RPA module fully focus on the information of current input $\bs{f}^{k,l-1}_{i}$.

\subsection{Optimized Searching for Unique Paths}

\subsubsection{Minimizing moving distance}
\begin{figure}[!t]
  \centering
  \includegraphics[width=\columnwidth]{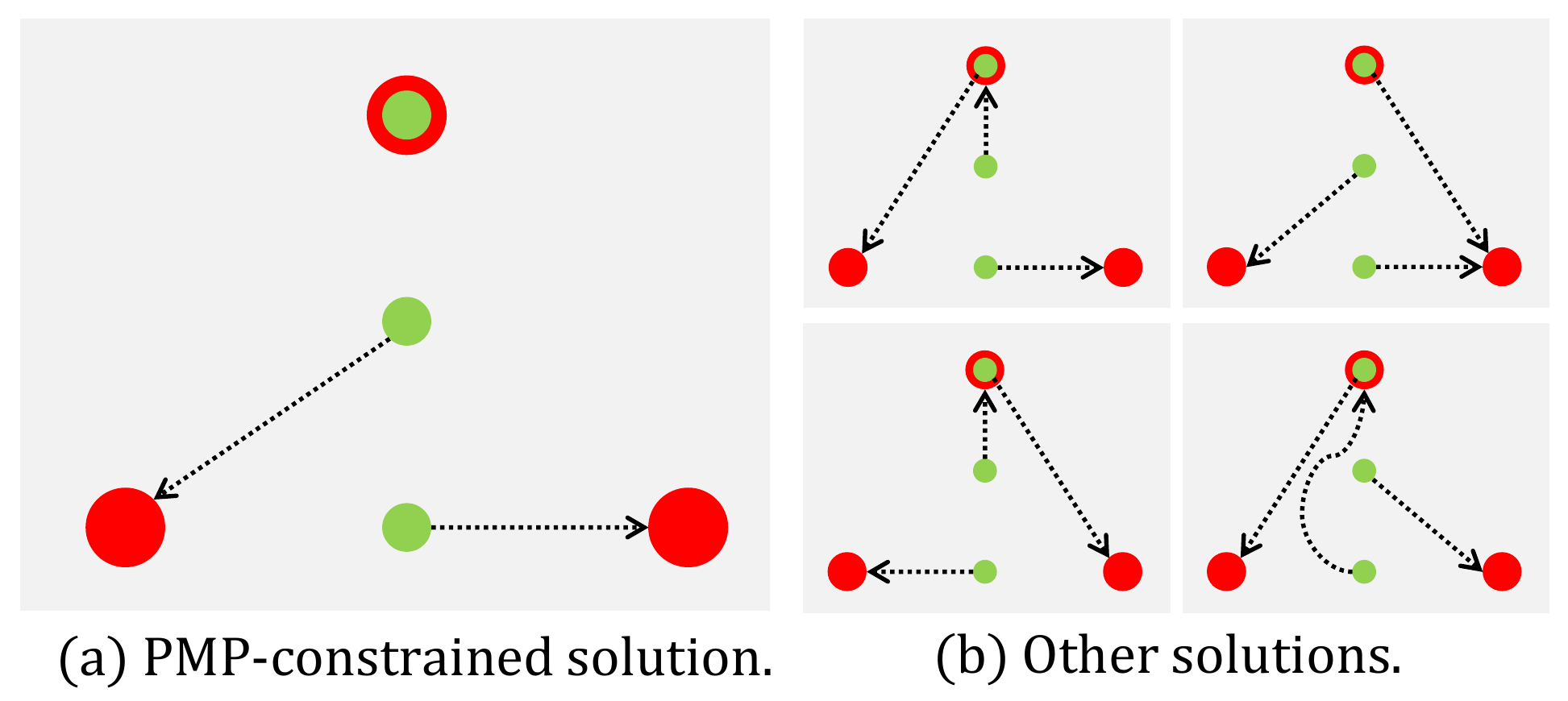}
  \caption{Illustration of multiple solutions when deforming input point cloud (green) into target point cloud (red). The PMD-constraint guarantees the uniqueness of point level correspondence (a) between input and target point cloud, and filter out various redundant solutions for moving points (b).
  }
  \label{fig:emd_solution}
\end{figure}

As shown in Figure \ref{fig:emd_solution}, the unordered nature of point cloud allows multiple solutions to deform the input shape into the target one, and the direct constraint (e.g. chamfer distance) on the deformed shape and its ground truth cannot guarantee the uniqueness of correspondence established between the input point set and the target point set. Otherwise, the network will be confused by the multiple solutions of point moving, which may lead to the failure of capturing detailed topology and structure relationships between incomplete shapes and complete ones.
In order to establish a unique and meaningful point-wise correspondence between input point cloud and target point cloud, we take the inspiration from Earth Mover's Distance \cite{rubner2000earth}, and propose to train PMP-Net++ to learn the path arrangement $\phi$ between source and target point clouds under the constraint of total point moving path distance. Specifically, given the source point clouds $\hat{X}=\{\hat{\bs{x}}_i|i=1,2,3,...,N\}$ and the target point cloud $X=\{\bs{x}_i|i=1,2,3,...,N\}$, we follow EMD to learn an arrangement $\phi$ which meets the constraint below
\begin{equation}\label{eq:emd}
  \mathcal{L}_{\mathrm{EMD}}(\hat{X},X)=\min_{\phi :\hat{X}\rightarrow X}\frac{1}{\hat{X}}\sum_{\hat{\bs{x}}\in\hat{X}}\| \hat{\bs{x}}-\phi(\hat{\bs{x}}) \|,
\end{equation}
In Eq.(\ref{eq:emd}), $\phi$ is considered as a bijection that minimizes the average distance between corresponding points in $\hat{X}$ and $X$.

Here, a unique correspondence between two point clouds can be guaranteed by the definition of EMD (Eq.(\ref{eq:emd})), which can be explained as follows: (1) deforming point cloud A into the shape of point cloud B (with the same number of points) equals to establish a bijection $\phi$ between point clouds A and B; (2) the bijection $\phi$ is unique when the sum of point displacement reaches the minimum value; (3) the optimization of Eq.(\ref{eq:emd}) is to minimize the sum of point displacement, which will yield a unique correspondence, following the data order determined by the input point cloud.

\begin{figure}[!t]
  \centering
  \includegraphics[width=\columnwidth]{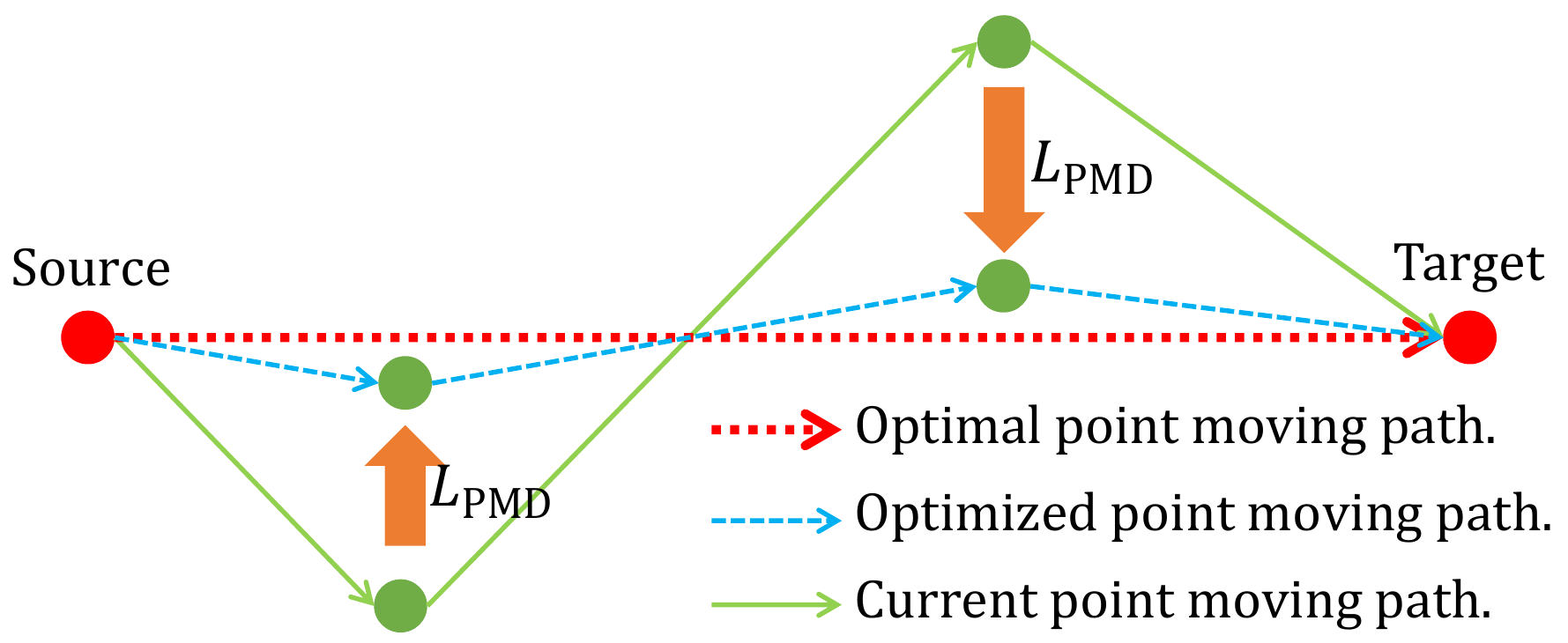}
  \caption{Illustration of the effectiveness of $\mathcal{L}_{PMD}$. By minimizing the point moving distance, the network is encouraged to learn more consistent paths from source to target, which will reduce redundant searching in each step and improve the efficiency.
  }
  \label{fig:mdl}
\end{figure}

According to Eq.(\ref{eq:emd}), bijection $\phi$ established by the network should achieve the minimum moving distance to move points from input shape to target shape. However, even if the correspondence between input and target point clouds is unique, there still exist various paths between source and target points, as shown in Figure \ref{fig:mdl}. Therefore, in order to encourage the network to learn an optimal point moving path, we choose to minimize the point moving distance loss ($\mathcal{L}_{\mathrm{PMD}}$), which is the sum of all displacement vector $\{\Delta \bs{p}_i^k\}$ output by all three steps in PMP-Net. 
The \emph{Point Moving Distance} loss is formulated as
\begin{equation}\label{eq:mdl}
  \mathcal{L}_{\mathrm{PMD}} = \sum_{k}\sum_{i}\|\Delta \bs{p}^k_i \|_2.
\end{equation}

Eq.(\ref{eq:mdl}) is more strict than EMD constraint. It requires not only the overall displacements of all point achieve the shortest distance, but also limits the point moving paths in each step to be the shortest one. Therefore, in each step, the network will be encouraged to search new path following the previous direction, as shown in Figure \ref{fig:mdl}, which will lead to less redundant moving decision and improve the searching efficiency.

\subsubsection{Multi-scaled searching radius}
PMP-Net++ searches the point moving path in a coarse-to-fine manner. For each step, PMP-Net++ reduces the maximum stride to move a point by the power of 10, which is, for step $k$, the displacement $\Delta\bs{p}^k_i$ calculated in Eq.(\ref{eq:displacement}) is limited to $10^{-k+1}\Delta\bs{p}^k_i$. This allows the network converges more quickly during training. And also, the reduced searching range will guarantee the network at next step not to overturn its decision made in the previous step, especially for the long range movements. Therefore, it can prevent the network from making redundant decision during path searching process.

\subsection{Extension to Dense Point Cloud Completion}
The deformation of point cloud cannot be directly used for increasing the number of points. Therefore, the input and the output point cloud must have the same resolution. Such characteristic of deformation based PMP-Net++ may become a problem when tackling 3D shapes with various number of points. In order to solve this problem, we propose to extend the PMP-Net++ to dense point cloud completion scenarios by adding noise to the input of each step. To achieve this goal, in step $k$, the input point cloud $\{\bs{p}_i^k\}$ is concatenated with a noise vector $\bs{\hat{n}}$ sampled from a standard normal distribution ${\rm N}(0,1)$ as:
\begin{equation}
  \{\bs{p}_i^k\}\leftarrow\{[\bs{p}_i^k:\bs{\hat{n}}]\}, \bs{\hat{n}}\sim{\rm N}(0,1)
\end{equation}
As a result, in each time, the point cloud varies from its previous data when input to the network. Then, by deforming point clouds multiple times and overlapping the deformation results together, we can get a dense point cloud with more points than the original input.

Note that different from the noise in Eq. \ref{eq:displacement}, which aims to push the points away from their original position in 3D space, the noise used in the input of each step aims to vary the final deformation results.

\noindent\textbf{More discussion.} The key idea of extension to dense point cloud completion is to increase the number of points. Such practice is commonly adopted in generation based methods. For PMP-Net, we duplicate points and concatenate with random noise to increase the number of points. And for the other methods like TopNet\cite{tchapmi2019topnet} and PCN \cite{yuan2018pcn}, they also need to duplicate features and concatenate with 2D grid code to increase points. The difference between PMP-Net++ and the other methods is that the duplication operation is settled at different stages in the network.

Note that if we simply repeat point coordinates, and move them without additional operations, the duplicated points will produce exactly the same displacement. Therefore, to solve this problem, in PMP-Net++, we add noise on each point features to force them moving to different places. Because the noise-enhanced input points have already been located at different spatial locations before the movement, these duplications will not be moved to the same target points.

\subsection{Training Loss}
The deformed shape is regularized by the complete ground truth point cloud through Chamfer distance (CD) and Earth Mover Distance (EMD). Following the same notations in Eq.(\ref{eq:emd}), the Chamfer distance is defined as:
\begin{equation}\label{eq:cd}
  \mathcal{L}_{CD}({X},\hat{X})=\sum_{\bs{x}\in X}\min_{\hat{\bs{x}}\in\hat{X}}\|\bs{x}-\hat{\bs{x}} \| + \sum_{\hat{\bs{x}}\in \hat{X}}\min_{\bs{x}\in X}\|\hat{\bs{x}}-\bs{x} \|.
\end{equation}
The total loss for training is then given as
\begin{equation}
  \mathcal{L}=\sum_{k}\mathcal{L}_{\mathrm{CD}}(P^k,P')+\mathcal{L}_{\mathrm{PMD}},
\end{equation}
where $P^k$ and $P'$ denote the point cloud output by step $k$ and the target complete point cloud, respectively. Note that finding the optimal $\phi$ is extremely computational expensive. In experiments, we follow the simplified algorithm in \cite{yuan2018pcn} to estimate an approximation of $\phi$.

\section{Experiments}
In this section, we first evaluate PMP-Net++ on general completion benchmark \emph{PCN} \cite{yuan2018pcn} and \emph{Completion3D} \cite{tchapmi2019topnet}. Then we further explore the potential application of PMP-Net++ on the point cloud up-sampling task. Finally, the effectiveness of each part of PMP-Net++ will be quantitatively evaluated through comprehensive ablation studies.
\subsection{Detailed Settings}
We use the single scale grouping (SSG) version of PointNet++ and its feature propagation module as the basic framework of PMP-Net. The detailed architecture of each part is described in Table \ref{table:encoder} and Table \ref{table:fp_module}, respectively.
\begin{table}[!h]
\centering
\caption{The detailed structure of encoder.}
\begin{tabular}{lcccc}
\toprule
Level &\#Points    &Radius   &\#Sample  &MLPs  \\ \midrule
1  &512  &0.2  &32  &$[64,64,128]$\\
2  &128  &0.4  &32  &$[128,128,256]$ \\
3  &-  &-  &- &$[256,512,1024]$ \\
\bottomrule
\end{tabular}
\label{table:encoder}
\end{table}

In Table \ref{table:encoder}, ``\#Points'' denotes the number of down-sampled points, ``Radius'' denotes the radius of ball query, ``\#Sample'' denotes the number of neighbors points sampled for each center point, ``MLPs'' denotes the number of output channels for MLPs in each level of encoder.

\begin{table}[!h]
\centering
\caption{The detailed architecture of feature propagation module.}
\begin{tabular}{lccc}
\toprule
Level &1 &2 &3  \\ \midrule
MLPs  &$[256,256]$ &$[256,128]$ &$[128,128,128]$\\
\bottomrule
\end{tabular}
\label{table:fp_module}
\end{table}

\begin{table*}[!t]
\centering
\caption{Point cloud completion on PCN dataset in terms of per-point L1 Chamfer distance $\times 10^{3}$ (lower is better).}
\begin{tabular}{l|c|cccccccc}
\toprule
Methods &Average  &Plane    &Cabinet  &Car   &Chair   &Lamp   &Couch    &Table    &Watercraft      \\ \midrule
FoldingNet  \cite{yang2018foldingnet}   &14.31  &9.49    &15.80    &12.61    &15.55   &16.41    &15.97    &13.65    &14.99   \\
TopNet  \cite{tchapmi2019topnet}     &12.15   &7.61   &13.31   &10.90    &13.82    &14.44      &14.78   &11.22  &11.12   \\
AtlasNet \cite{groueix2018atlasnet}   &10.85   &6.37    &11.94    &10.10    &12.06    &12.37    &12.99    &10.33    &10.61   \\
PCN  \cite{yuan2018pcn}   &9.64 &5.50    &22.70    &10.63    &8.70    &11.00    &11.34    &11.68    &8.59   \\
GRNet \cite{xie2020grnet}  &8.83   &6.45   &10.37   &9.45    &9.41    &7.96      &10.5   &8.44  &8.04   \\
CRN \cite{wang2020cascaded}  &8.51   &4.79   &9.97   &\textbf{8.31}    &9.49    &8.94      &10.69   &7.81  &8.05   \\
NSFA \cite{zhang2020detail}  &8.06 &4.76 &10.18 &8.63 &8.53 &7.03 &10.53 &7.35 &7.48 \\
PMP-Net \cite{wen2021pmp} &8.66  &5.50    &11.10    &9.62    &9.47    &6.89    &10.74    &8.77   &7.19 \\\midrule
PMP-Net++(no PMD, Ours) &7.74  &4.69   &10.15    &8.78    &8.30    &6.08    &10.28    &7.23  &6.61 \\
PMP-Net++(Ours) &\textbf{7.56}  &\textbf{4.39}   &\textbf{9.96}    &8.53    &\textbf{8.09}    &\textbf{6.06}    &\textbf{9.82}    &\textbf{7.17}  &\textbf{6.52} \\
\bottomrule
\end{tabular}
\label{table:PCN}
\end{table*}
We use AdamOptimizer to train PMP-Net++ with an initial learning rate $10^{-3}$, and exponentially decay it by 0.5 for every 20 epochs. The training process is accomplished using a single NVDIA GTX 2080TI GPU with a batch size of 24. PMP-Net++ takes 150 epochs to converge on both PCN and Completion3D dataset. We scale all input training shapes of Completion3D by 0.9 to avoid points out of the range of \emph{tanh} activation.

Note that currently the transformer is only involved in encoder, and we do not observe significant performance gain when adding transformer in decoder network. Since the key operation of feature propagation module (FP-module in decoder) is the trilinear interpolation (which is not learnable), the quality of point features produced by the decoder are mainly relies on the encoder. Therefore, additional transformer in decoder may not help the network too much to enhance the quality of point features.

\begin{figure*}[!t]
  \centering
  \includegraphics[width=0.95\textwidth]{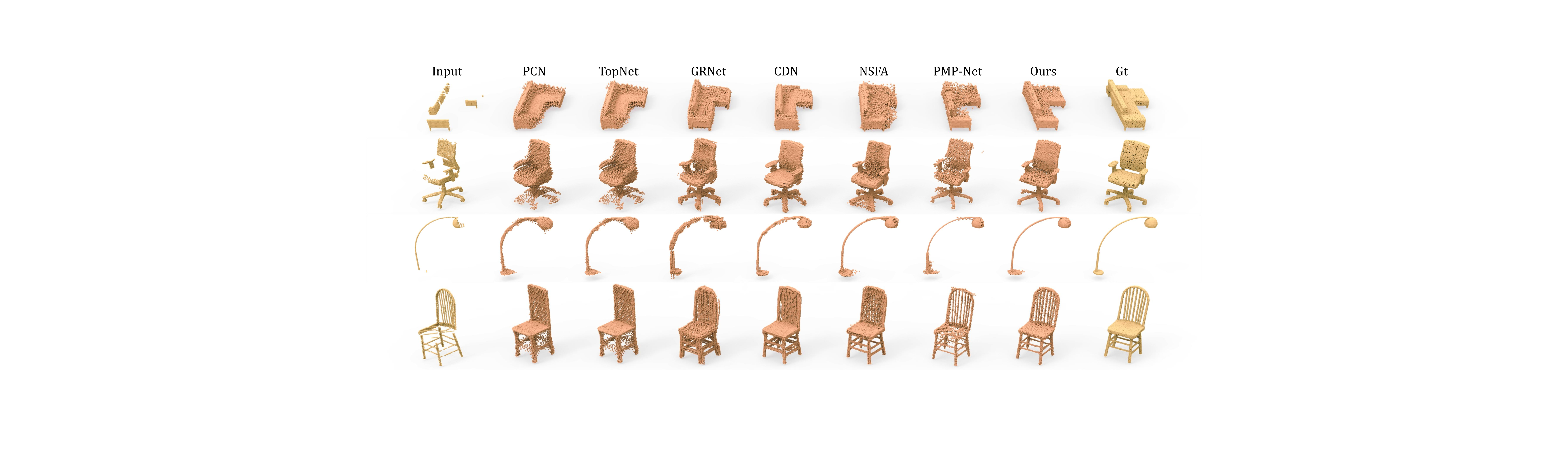}
  \caption{Visualization of point cloud completion comparison with previous methods on PCN dataset.
  }
  \label{fig:pcn_comp}
\end{figure*}

\begin{figure*}[!t]
  \centering
  \includegraphics[width=\textwidth]{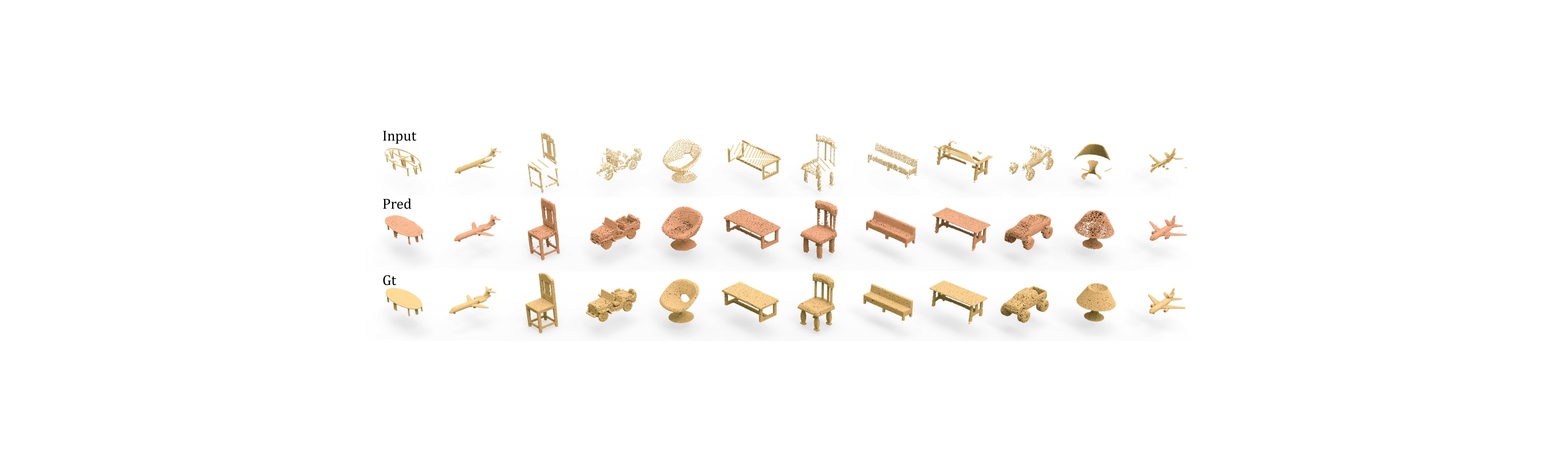}
  \caption{Visualization of more completion results using our PMP-Net++ on PCN dataset.
  }
  \label{fig:pcn_vis}
\end{figure*}

\subsection{Point Cloud Completion on PCN Dataset}
\subsubsection{Dataset and evaluation metric}
We show that PMP-Net++ learned on sparse point cloud can be directly applied to the dense point cloud completion. Specifically, we keep training PMP-Net++ on sparse shape with 2,048 points, and reveal its generalization ability by predicting dense complete shape with 16,384 points on PCN dataset \cite{yuan2018pcn}. PCN dataset is derived from ShapeNet dataset, in which each complete shape contains 16,384 points. The partial shapes have various point numbers, so we first down-sample shapes with more than 2,048 points to 2,048, and up-sample shapes with less than 2048 points to 2048 by randomly copying points.
Since PMP-Net++ learns to move points instead of generating points, it requires the same number of points in incomplete point cloud and complete one. In order to predict complete shape of 16,384 points, we repeat 8 times of prediction for each shape during testing, with each time moving 2,048 points. Note that PMP-Net++ is still trained on sparse point clouds with 2,048 points, which are sampled from the dense point clouds of PCN dataset.

On PCN dataset, we use the per-point L1 Chamfer distance (CD) as the evaluation metric, which is the CD in Eq.(\ref{eq:cd}) averaged by the point number.

\subsubsection{Quantitative comparison}
The comparison in Table \ref{table:PCN} shows that PMP-Net++ yields a comparable performance to the state-of-the-art method \cite{wang2020cascaded}, and ranks first on PCN dataset. The result of Wang et al.\cite{wang2020cascaded} is cited from its original paper, while the results of other compared methods are all cited from \cite{xie2020grnet}. Note that most generation based methods (like Wang et al. \cite{wang2020cascaded} and GRNet \cite{xie2020grnet} in Table \ref{table:PCN}) specially designed a coarse-to-fine generation process in order to obtain better performance on dense point cloud completion. In contrast, our PMP-Net++ trained on 2,048 points can directly generate arbitrary number of dense points by simply repeating the point moving process, and still achieves comparable results to the counterpart methods. Moreover, we further discuss the effectiveness of PMD loss by comparing the baseline PMP-Net++ with \emph{no PMD} variation, which we remove the PMD loss during training. From Table \ref{table:PCN}, we can find that PMD loss effectively improves the performance of PMP-Net++, which is in accordance with our opinion that point moving path should be regularized to better capture the detailed topology and structure of 3D shapes.

\subsubsection{Qualitative comparison}
In Figure \ref{fig:pcn_comp} and Figure \ref{fig:pcn_vis}, we further demonstrate the advantage of PMP-Net++ over other methods, by visually compare the completion results on PCN dataset. Specifically, in Figure \ref{fig:pcn_comp}, we compare our PMP-Net++ with other counterparts cross different object categories. For example, in the third row of Figure \ref{fig:pcn_comp}, the task is to predict the complete shape of an incomplete lamp. In this task, most of the evaluated methods failed to preserve the detailed geometries of the lamppost, where many noise points are distributed around the lamppost. Compared with the generative methods in Figure \ref{fig:pcn_comp}, our PMP-Net++ can reveal a high quality lamppost. Moreover, when comparing the PMP-Net++ with PMP-Net, we can find that PMP-Net++ achieves a better shape prediction at the bottom and the top of the lamppost. The advantages of inferring and preserving detailed shapes of PMP-Net++ can also be well proved by the observation of the fourth row: by comparing the reconstruction results of the chair back, we can find that PMP-Net++ can clearly preserve the detailed shapes of each beam on the chairback; by comparing the completion results of the chair legs, the results of PMP-Net++ is also closer to the ground truth than its any other counterparts. Moreover, the comparison of the sofa between the PMP-Net and PMP-Net++ visually reveals the improvements of PMP-Net++ over the PMP-Net. The complete sofa predicted by PMP-Net distributes less points in the missing region of the sofa than the PMP-Net++.

\begin{table*}[!t]
\centering
\caption{Point cloud completion on Completion3D dataset in terms of per-point L2 Chamfer distance $\times 10^{4}$ (lower is better).}
\begin{tabular}{l|c|cccccccc}
\toprule
Methods &Average  &Plane    &Cabinet  &Car   &Chair   &Lamp   &Couch    &Table    &Watercraft      \\ \midrule
FoldingNet  \cite{yang2018foldingnet}   &19.07  &12.83    &23.01    &14.88    &25.69    &21.79    &21.31    &20.71    &11.51   \\
PCN  \cite{yuan2018pcn}   &18.22 &9.79    &22.70    &12.43    &25.14    &22.72    &20.26    &20.27    &11.73   \\
PointSetVoting \cite{zhang2021point}  &18.18 &6.88    &21.18    &15.78    &22.54    &18.78    &28.39    &19.96    &11.16   \\
AtlasNet \cite{groueix2018atlasnet}   &17.77   &10.36    &23.40    &13.40    &24.16    &20.24    &20.82    &17.52    &11.62   \\
SoftPoolNet \cite{wang2020softpoolnet} &16.15   &5.81   &24.53   &11.35    &23.63    &18.54      &20.34   &16.89  &7.14   \\
TopNet  \cite{tchapmi2019topnet}     &14.25   &7.32   &18.77   &12.88    &19.82    &14.60      &16.29   &14.89  &8.82   \\
SA-Net \cite{wen2020sa}  &11.22   &5.27   &14.45   &7.78    &13.67    &13.53      &14.22   &11.75  &8.84   \\
GRNet \cite{xie2020grnet}  &10.64   &6.13   &16.90   &8.27    &12.23    &10.22      &14.93   &10.08  &5.86   \\
CRN \cite{wang2020cascaded} &9.21	&3.38	&13.17	&8.31	&10.62	&10.00	&12.86	&9.16	&5.80\\
PMP-Net \cite{wen2021pmp} &9.23  &3.99    &14.70    &8.55    &10.21    &9.27    &12.43    &8.51   &5.77 \\
VRCNet \cite{pan2021vrc} &8.12    &3.94   &13.46  &\textbf{6.72}   &10.35  &9.87   &12.48  &7.73   &6.14 \\\midrule
PMP-Net++(Ours) &\textbf{7.97}  &\textbf{3.25}   &\textbf{12.25}    &7.62    &\textbf{8.71}    &\textbf{7.64}    &\textbf{11.6}    &\textbf{7.06}  &\textbf{5.38} \\
\bottomrule
\end{tabular}
\label{table:completion3D}
\end{table*}

\begin{figure*}[!t]
  \centering
  \includegraphics[width=0.95\textwidth]{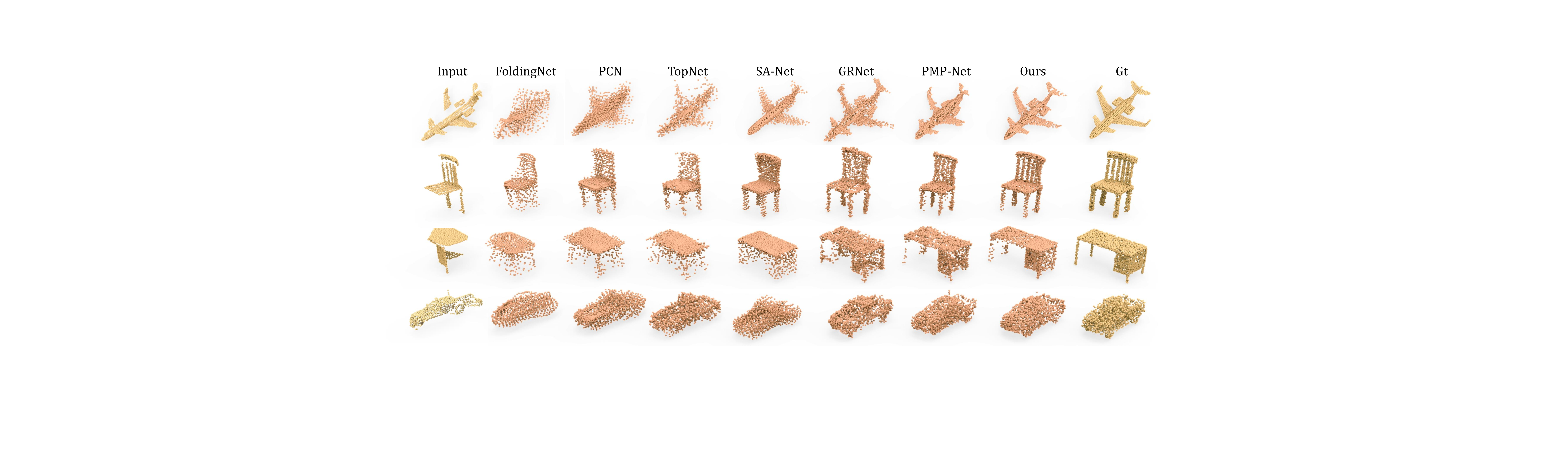}
  \caption{Visualization of point cloud completion comparison with previous methods on Completion3D dataset.
  }
  \label{fig:completion3d_comp}
\end{figure*}

\subsection{Point Cloud Completion on Completion3D Dataset}
\subsubsection{Dataset and evaluation metric}
We evaluate our PMP-Net++ on the widely used benchmark of 3D point cloud completion, i.e. \emph{Completion3D} \cite{tchapmi2019topnet}, which is a large-scaled 3D object dataset derived from the ShapeNet dataset. The partial 3D shapes are generated by back-projecting 2.5D depth images from partial views into 3D space. Completion3D dataset concerns the completion task of sparse point cloud completion, where it generates only one partial view for each complete 3D object in ShapeNet dataset, and samples 2,048 points from the mesh surface for both the complete and partial shapes.
We follow the settings of training/validation/test splits in Completion3D for fair comparison with the other methods.

Following previous studies \cite{tchapmi2019topnet,xie2020grnet,wen2020sa,yuan2018pcn}, we use the per-point L2 Chamfer distance (CD) as the evaluation metric on Completion3D dataset. L2 Chamfer distance is the CD in Eq.(\ref{eq:cd}) averaged by the point number, where the L1-norm in Eq.(\ref{eq:cd}) is replaced by L2-norm.

\begin{figure*}[!t]
  \centering
  \includegraphics[width=0.95\textwidth]{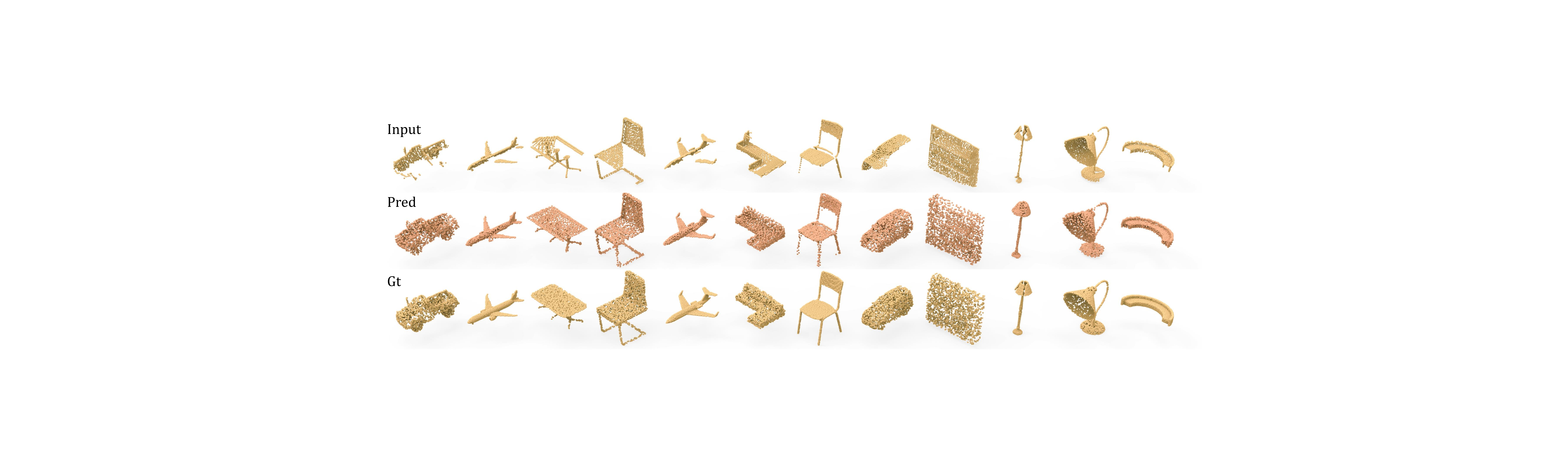}
  \caption{More completion results using our PMP-Net++ on Completion3D dataset.
  }
  \label{fig:completion3d_self}
\end{figure*}

\subsubsection{Quantitative comparison}
The quantitative comparison results\footnote{Results are cited from \url{https://completion3d.stanford.edu/results}} of PMP-Net++ with the other state-of-the-art point cloud completion methods are shown in Table \ref{table:completion3D}, in which the PMP-Net++ achieves the best performance in terms of average chamfer distance across all categories. The second best published method on Completion3D leaderboard  is VRCNet\cite{pan2021vrc}, which achieves 8.12 in terms of average CD, and PMP-Net++ improves such state-of-the-art performance by 0.17 (also in terms of average CD). When considering per-category performance, PMP-Net++ achieves the best results in 7 out of 8 categories across all compared counterpart methods, which justifies the better generalization ability of PMP-Net++ across different shape categories. Note that, compared with PMP-Net, PMP-Net++ significantly reduces the average CD loss on Completion3D dataset by 13.8\%, and outperforms the PMP-Net on all 8 categories in terms of per-category CD.
As we discussed in Section \ref{sec:rel}, GRNet \cite{xie2020grnet} is a voxel aided shape completion method, where the completion process utilizes information from two different data modalities (i.e point cloud and voxels). The better performance of PMP-Net++ over GRNet proves the effectiveness of our method, which can exploit the abundant geometric information lying in the point clouds. Other methods like SA-Net \cite{wen2020sa} in Table \ref{table:completion3D} are typical generative completion methods which are fully based on point clouds, and the nontrivial improvement of PMP-Net++ over these methods justifies the effectiveness of deformation based solution in point cloud completion task.

\begin{figure*}[!t]
  \centering
  \includegraphics[width=\textwidth]{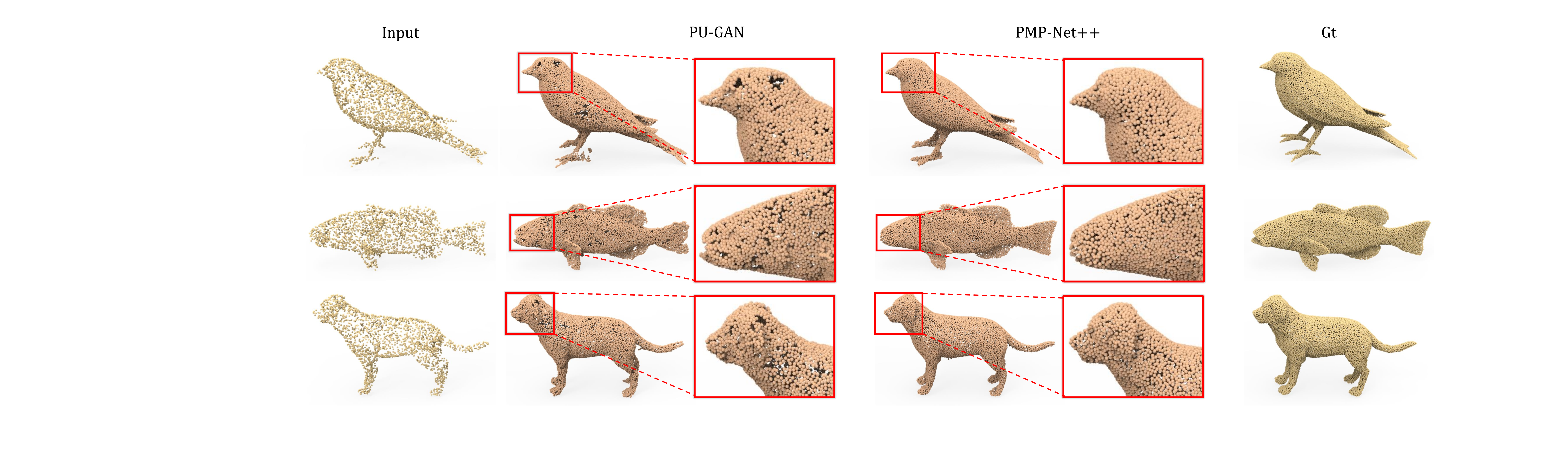}\vspace{-0.4cm}
  \caption{Visualization of point cloud up-sampling. We typically compare our PMP-Net++ with the state-of-the-art up-sampling method PU-GAN \cite{li2019pu}, and demonstrate the advantages of PMP-Net++ for generating better detailed shapes.
  }
  \label{fig:upsample_comp}
\end{figure*}

\begin{figure}[!t]
  \centering
  \includegraphics[width=\columnwidth]{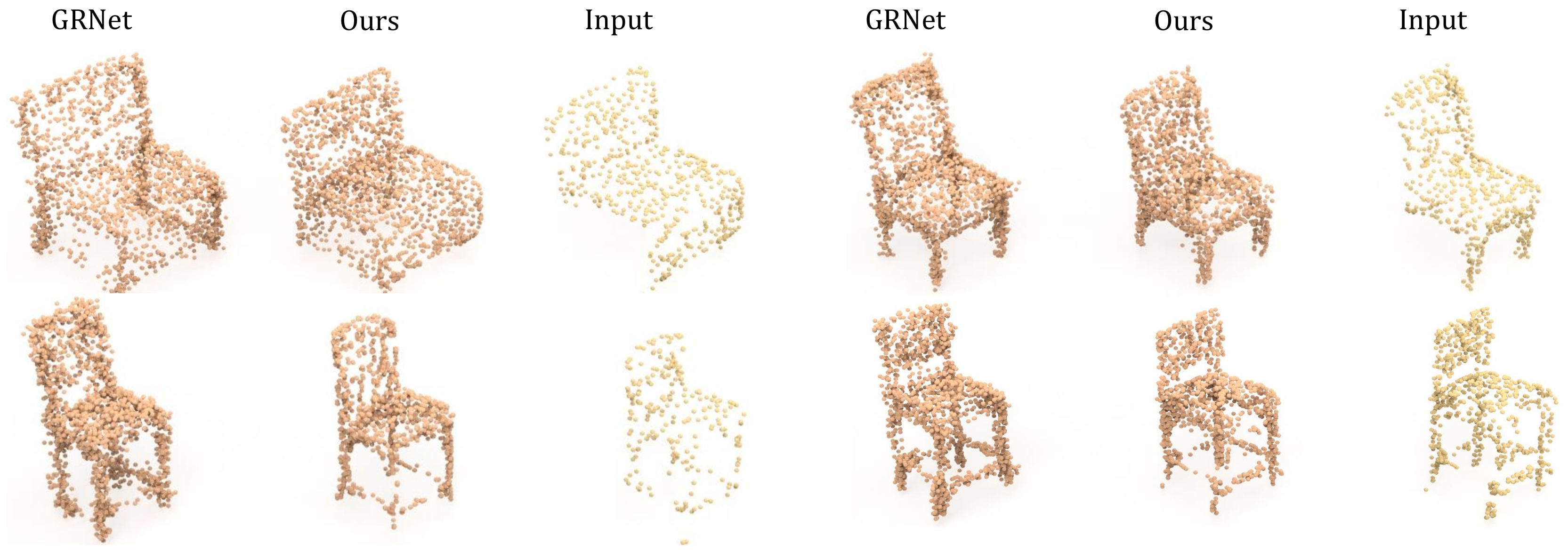}\vspace{-0.4cm}
  \caption{Visual comparison of PMP-Net++ and GRNet on ScanNet chairs.
  }
  \label{fig:scannet_vis}
\end{figure}

\subsubsection{Qualitative comparison}
In Figure \ref{fig:completion3d_comp}, we visually compare PMP-Net++ with the other completion methods on Completion3D dataset, from which we can find that PMP-Net++ predicts much more accurate complete shapes on various shape categories, while other methods may output some failure cases for certain input shapes. For example, the input \emph{table} in Figure \ref{fig:completion3d_comp} (the third row) loses half of its legs and surface. The completion results made by GRNet and PMP-Net~ almost predict the right overall shape of the complete desktop, but fail to reconstruct the detailed structure like clean legs and smooth surface. On the other hand, methods like FoldingNet \cite{yang2018foldingnet}, PCN \cite{yuan2018pcn}, TopNet \cite{tchapmi2019topnet} and SA-Net \cite{wen2020sa} intend to repair the desktop but fail to predict a complete overall shape. Moreover, the advantage of deformation based PMP-Net++ over the generative methods can be well proved by the case of \emph{chair} in Figure \ref{fig:completion3d_comp} (the second row). Generative methods, especially like GRNet, successfully learn the complete structure of the input chair, but fail to reconstruct the columns on the chair back, which is the residual part of input shape. On the other hand, the deformation based PMP-Net++ can directly preserve the input shape by moving just a small amount of point to perform completion on certain areas, and keep the input shape unchanged. In Figure \ref{fig:completion3d_self}, we further visualize more completion results of PMP-Net.

\subsubsection{Extension to ScanNet chairs}
To evaluate the generalization ability of PMP-Net++ on point cloud completion task, we pre-train PMP-Net++ on Completion3D dataset and evaluate its performance on the chair instances in ScanNet dataset without finetuning, and compare with GRNet (which is the second best method in Table \ref{table:completion3D}, and also pre-trained on Completion3D). The visual comparison is shown in Figure \ref{fig:scannet_vis}. Since there is no ground truth for ScanNet dataset, we typically follow \cite{yuan2018pcn} to report partial metrics (Fidelity and MMD) in Table \ref{table:scannet_chair} based on the selected chairs of Figure \ref{fig:scannet_vis}. The PMP-Net++ completes shapes with less noise than GRNet, which benefits from its point moving practice. Because the differences in data distribution between Completion3D and ScanNet will inevitably confuse the network, and the point moving based PMP-Net++ can simply choose to leave those points in residual part of an object to stay at their own place to preserve a better shape, in contrast, generation based GRNet has to predict new points for both residual and missing part of an object.

\begin{table}[!h]
\centering
\caption{Quantitative evaluation of ScanNet chairs.}
\begin{tabular}{lcc}
\toprule
Methods &Fidelity($10^{4}$)    &MMD($10^{3}$)  \\ \midrule
GRNet \cite{xie2020grnet}  &2.95  &3.17 \\
PMP-Net++   &2.05  &2.65  \\
\bottomrule
\end{tabular}
\label{table:scannet_chair}
\end{table}

\subsection{Point Cloud Up-sampling}
\subsubsection{Dataset and evaluation metric}
In this section, we experimentally demonstrate the effectiveness of PMP-Net++ on other similar task, i.e. the point cloud up-sampling task. The scenario encountered by PMP-Net++ on this task is the same as the dense point cloud completion, where the difficulty is to increase the number of points in order to reveal more detailed geometric information of 3D model. Therefore, we adopt the dense completion version of PMP-Net++ to the up-sampling task. For fair comparison with the previous counterpart methods, we follow the same practice of PU-GAN to use its dataset and experimental settings for evaluation. According to PU-GAN, the dataset is a collection of 147 models from the dataset of PU-Net, MPU and Vision-air repository, where 120 models are selected for training and the rest 27 models are used for testing. The commonly used Chamfer distance (CD) and Hausdorff distance are adopted as our evaluation metric.

\subsubsection{Quantitative and qualitative comparison}
The quantitative comparison is given in Table \ref{table:up-sampling}, from which we can find that our PMP-Net++ achieves the best performance among the compared counterparts. Note that, in point cloud up-sampling task, we use exact the same network settings and structures as the point cloud completion on PCN dataset. Therefore, the better results on point cloud up-sampling of our network proves the generalization ability of PMP-Net++ to various tasks.
\begin{table}[!h]
\centering
\caption{Quantitative comparison on point cloud up-sampling task.}
\begin{tabular}{lcc}
\toprule
Methods &CD($10^{-3}$)    &HD($10^{-3}$)  \\ \midrule
EAR \cite{huang2013edge}  &0.52  &7.37 \\
PU-Net \cite{li2018pu}  &0.72  &8.94  \\
MPU \cite{yifan2019patch} &0.49  &6.11  \\
PU-GAN \cite{li2019pu}  &0.28  &4.64  \\ \midrule
PMP-Net++(Ours) &\textbf{0.26} &\textbf{4.63} \\
\bottomrule
\end{tabular}
\label{table:up-sampling}
\end{table}

In Figure \ref{fig:upsample_comp}, we further illustrate the better performance of PMP-Net++ by visually compare our network with PU-GAN. For example, in the completion of the head of the bird (first row of Figure \ref{fig:upsample_comp}), the up-sampling results of PMP-Net++ distribute the points more evenly than the PU-GAN. From the results of PU-GAN, we can still observe several holes on the bird's head, while on the result of PMP-Net++, there is no such incomplete region appearing on the surface of the bird's head. Moreover, the up-sampled points using PMP-Net++ are distributed more evenly compared with the PU-GAN, especially on the head of the fish in the second row of Figure \ref{fig:upsample_comp}.

\subsection{Model Analysis}
In this subsection, we analyze the influence of different parts in the PMP-Net++, and compared it with PMP-Net to analysis the effectiveness of the incremental contribution. By default, we use the same network sittings in PMP-Net\cite{wen2021pmp} for all experiments, where all studies are typically conducted on the validation set of Completion3D dataset under four categories (i.e. plane, car, chair and table) for convenience.

\subsubsection{Analysis of RPA module and PMP loss}
We analyze the effectiveness of RPA module by replacing it with other units in PMP-Net++. And for PMP loss, we analyze its effectiveness by removing PMP loss from the network. Specifically, we develop six different variations for comparison: (1) \emph{NoPath} is the variation that removes the RPA module from the network; (2) \emph{Add} is the variation that replaces RPA module with element-wise add layer in the network; (3) RNN, (4) LSTM and (5) GRU are variations that replace RPA module with different recurrent unit.

The shape completion results are shown in Table \ref{table:analysis_rpa}, in which we report the results of both PMP-Net and PMP-Net++ for a comprehensive comparison. From Table \ref{table:analysis_rpa}, we can find that two baseline variations (which uses the RPA module) achieve the best performance on two backbones, respectively, which justifies the effectiveness of the proposed RPA module across different network structure. The worst result across different kinds of unit is yielded by Add variation under the PMP-Net backbone, while under the PMP-Net++ backbone, the worst result is yielded by the RNN unit. The different performance of Add variation and GRU variation cross different backbones indicates that, recurrent network structures cannot provide a robust aggregation of sequential information, which is generated during the path searching process of PMP-Net++ framework. Moreover, since the RPA module is originated from the GRU unit, the comparisons between RPA baseline and GRU variation on two backbones justify the effectiveness of our designation of RPA module, which can give more consideration to the information from current step than GRU unit, and help the network to make more precise decision for point moving.

The effectiveness of newly added transformer unit can be fully justified by the comparison between the backbones of PMP-Net and PMP-Net++, where the best performance with the PMP-Net backbones is still worse than the worst performance of PMP-Net++, no matter which kind of variation they use.
\begin{table}[!h]
\centering
\caption{Analysis of RPA and PMP loss (baseline marked by ``*'').}
\begin{tabular}{l|lccccc}
\toprule
Backbone& Unit. &avg.    &plane   &chair  &car  &table  \\ \midrule
\multirow{7}{*}{PMP-Net} &NoPath &11.95 &3.55 &8.30  &16.15 &19.79\\
&Add  &12.23  &\textbf{3.32}    &16.47 &8.10 &21.05\\
&RNN  &12.12  &3.55   &16.19 &8.14  &20.58\\
&LSTM  &11.99  &3.79  &15.37  &8.09  &20.72\\
&GRU  &11.87  &3.44  &15.44 &7.85  &20.72\\
&baseline* &\textbf{11.58}  &3.42   &\textbf{15.88} &\textbf{7.87} &\textbf{19.15}\\\midrule
\multirow{7}{*}{PMP-Net++}& NoPath &8.09 &2.47 &10.8  &6.65 &12.31\\
&Add  &8.07  &\textbf{2.39}  &10.90  &6.54 &12.40\\
&RNN  &8.54  &2.46  &11.70  &6.86 &13.00\\
&LSTM  &8.30  &2.47  &10.92  &6.69 &13.1\\
&GRU  &8.27  &2.40  &11.31  &6.63 &12.62\\
&baseline* &\textbf{8.03}  &2.58  &\textbf{10.70}  &\textbf{6.44} &\textbf{12.30}\\
\bottomrule
\end{tabular}
\label{table:analysis_rpa}
\end{table}

\subsubsection{Effect of multi-step path searching}
In Table \ref{table:step_analysis}, we analyze the effect of different steps for point cloud deformation, and we also compare the performance between PMP-Net and PMP-Net++. Specifically, the ratio of searching radius between each step is fixed to $10$, and then, the number of steps to deform the point clouds is set to 1, 2 and 4, respectively. For example, when the step is set to 4, the corresponding searching radius is $\{1.0, 10^{-1}, 10^{-2}, 10^{-3}\}$. And the searching radius for step=2 is set to $\{1.0, 10^{-1}\}$. By comparing the results of step 1, 2 and 3 from Table \ref{table:step_analysis}, we can find that under both backbones, deforming point cloud by multiple steps effectively improves the completion performance. On the other hand, the comparison between step 3 and step 4 shows that the performance of multi-step path searching will reach its limitation, because too many steps may cause information redundancy in path searching.
It can also be noticed that the gap between 1 step v.s. 2 step for PMP-Net++ (9.45 v.s. 8.36) is much larger than PMP-Net (12.26 v.s. 11.90). This can be dedicated to the point transformer enhanced module, which enables PMP-Net++ to predict more accurate shape completion than PMP-Net. Therefore, based on the output of step 1, where PMP-Net++ yields a CD of 9.45 and PMP-Net is 12.26 according to Table \ref{table:step_analysis}, at step 2, the point transformer enhanced module of PMP-Net++ can learn more efficient geometric information, which achieves better performance (9.45 vs. 8.36) than PMP-Net (12.26 vs. 11.90).

\begin{table}[!h]
\centering
\caption{The effect of different steps (baseline marked by ``*'').}
\begin{tabular}{l|lccccc}
\toprule
Backbone&Steps. &avg.    &plane   &chair  &car  &table  \\ \midrule
\multirow{4}{*}{PMP-Net} &1  &12.26  &3.71   &15.59 &8.27  &21.48\\
&2  &11.90  &3.47    &15.66 &7.95 &20.53\\
&3*  &\textbf{11.58}  &3.42    &\textbf{15.88} &\textbf{7.87} &\textbf{19.15}\\
&4  &11.67  &\textbf{3.39}    &15.89 &7.91 &19.48\\\midrule
\multirow{4}{*}{PMP-Net++} &1  &9.45  &2.64  &12.90  &7.26 &14.90\\
&2  &8.36  &2.61  &11.63  &6.66 &12.50\\
&3*  &\textbf{8.03}  &2.58  &10.70  &\textbf{6.44} &\textbf{12.30}\\
&4  &8.05  &\textbf{2.32}  &\textbf{10.61}  &6.51 &12.60\\
\bottomrule
\end{tabular}
\label{table:step_analysis}
\end{table}

\subsubsection{Visual analysis of multi-step searching}
\begin{figure*}[!t]
  \centering
  \includegraphics[width=\textwidth]{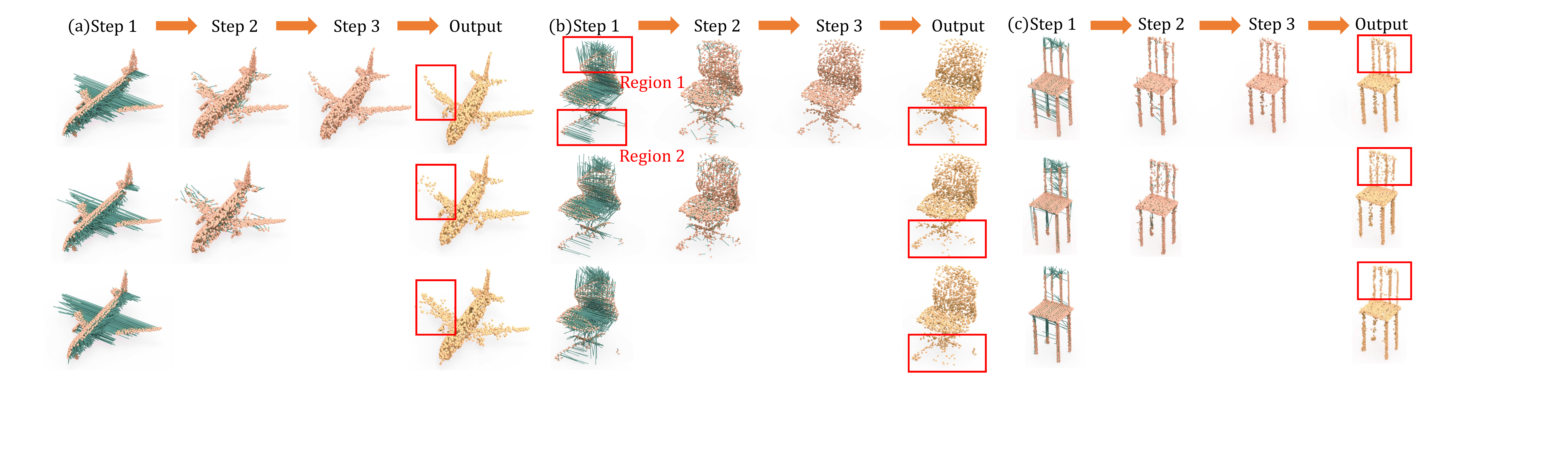}\vspace{-0.4cm}
  \caption{Illustration of multi-step searching under different searching steps. The first row is 3-step completion, and the second row is 2-step completion, and so on. The 4-step completion have the similar visual effects  as 3-step completion. Due to very short searching radius for visualizing the displacement at 4-th step, we only illustrate step 1, 2 and 3 in this figure.
  }
  \label{fig:step_analysis}
\end{figure*}

We visualize the point deformation process under different searching step sittings in Figure \ref{fig:step_analysis}. Comparing the 3-step searching in the top-row with the other two sittings, the empty space on the chair back is shaped cleaner as highlighted by rectangles, which justifies the effectiveness of multi-step searching to consistently refine the shape.
 Moreover, from the visualization of Figure \ref{fig:step_analysis}, we also find that the network can actually use some edge information to infer the destination of moving point cloud. Take the chair in Figure \ref{fig:step_analysis}(b) as an example, we can draw two conclusions as follows.
 \begin{itemize}
     \item In Region 1 of Figure \ref{fig:step_analysis}(b), the network moves the points on the chair back to complete the missing upper part, where the points on the edge of chair back are moved above, because they are geometrically closer to the missing part. Such point moving pattern utilizes the edge information of incomplete shape.
     \item In Region 2 of Figure \ref{fig:step_analysis}(b), the network moves the point from another chair legs to complete the missing legs. We can find that the point moving paths are almost parallel to each other, where each point on one leg is moved to the same place on the other leg. The point moving pattern in Region 2 clearly follows the one-to-one geometric correspondence between the points on two legs.
\end{itemize}
In all, the network not only learns the one-to-one geometric correspondence between the points, but also takes the geometric distance between the border of incomplete shape and its missing part into account during completion. Considering that the PMD loss focuses on regularizing the geometric correspondence, and there is no constraints for learning the border information, the improvements of PMP-Net++ mainly comes from the geometric correspondence.

\subsubsection{The shape to apply deformation}

\begin{figure}[!h]
  \centering
  \includegraphics[width=\columnwidth]{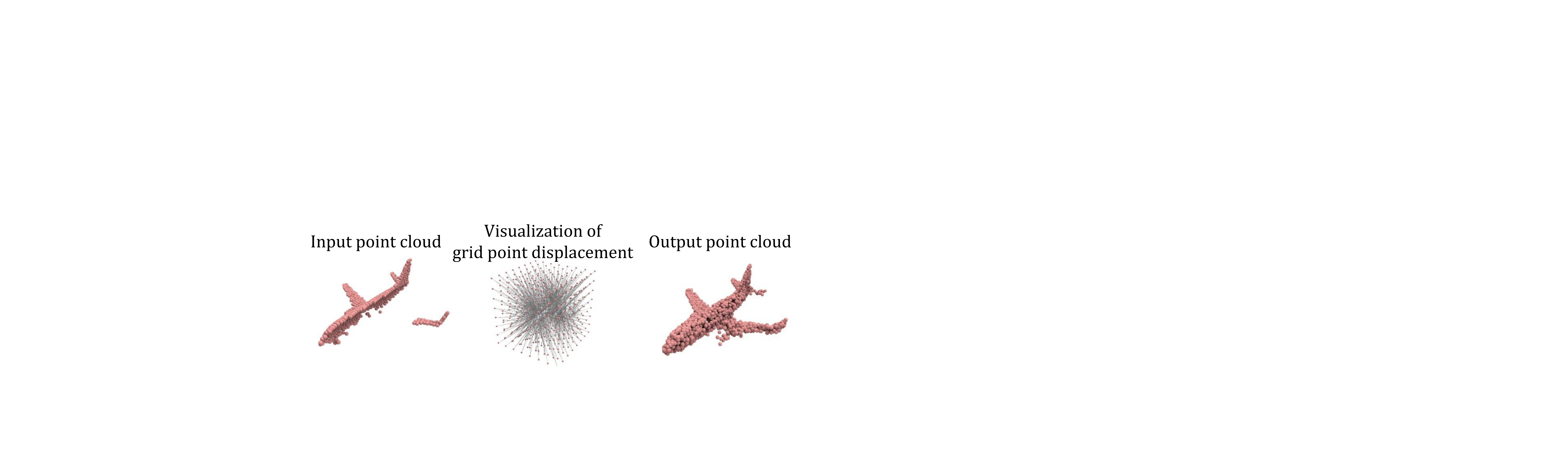}\vspace{-0.4cm}
  \caption{Illustration of deformation from grid point of size $8\times 8\times 8$.
  }
  \label{fig:grid_analysis}
\end{figure}

Deforming shapes from incomplete point cloud is not the only choice for PMP-Net++. Therefore, we provide and discuss one more possible choice of shape deformation, which is to deform 512 grid points (aranged as $8\times 8\times 8$ cubic) into a complete shape. Specifically, we use one-step version of PMP-Net++ for convenience, and use trilinear interpolation to propagate point cloud features onto these grid points. In order to generate 2048 points, we duplicate each grid point 4 times and concatenate each duplication with a random noise. Thus, each grid point will be encouraged to be split into 4 sub-points, and the all 512 grid points will finally output 2048 points. The performance and visualization are given in Table \ref{table:grid_analysis} and Figure \ref{fig:grid_analysis}, respectively.

Table \ref{table:grid_analysis} shows that grid points yield 9.98 of the average CD, which is lower but relatively comparable to PMP-Net++. We believe the effectiveness of grid points comes from the ordered data predictions, and the better performance of PMP-Net++ comes from utilizing more geometric information, which is provided by the deformation process from the incomplete point cloud. 
\begin{table}[!h]
\centering
\caption{Comparison with deformation from grid points.}
\begin{tabular}{l|lccccc}
\toprule
Methods. &avg.    &plane   &chair  &car  &table  \\ \midrule
PMP-Net++(one step) &9.45  &2.64  &12.90   &7.26 &14.90 \\
Grid &9.98  &3.34  &13.71    &7.45 &15.42\\
\bottomrule
\end{tabular}
\label{table:grid_analysis}
\end{table}

\begin{figure*}[!t]
  \centering
  \includegraphics[width=\textwidth]{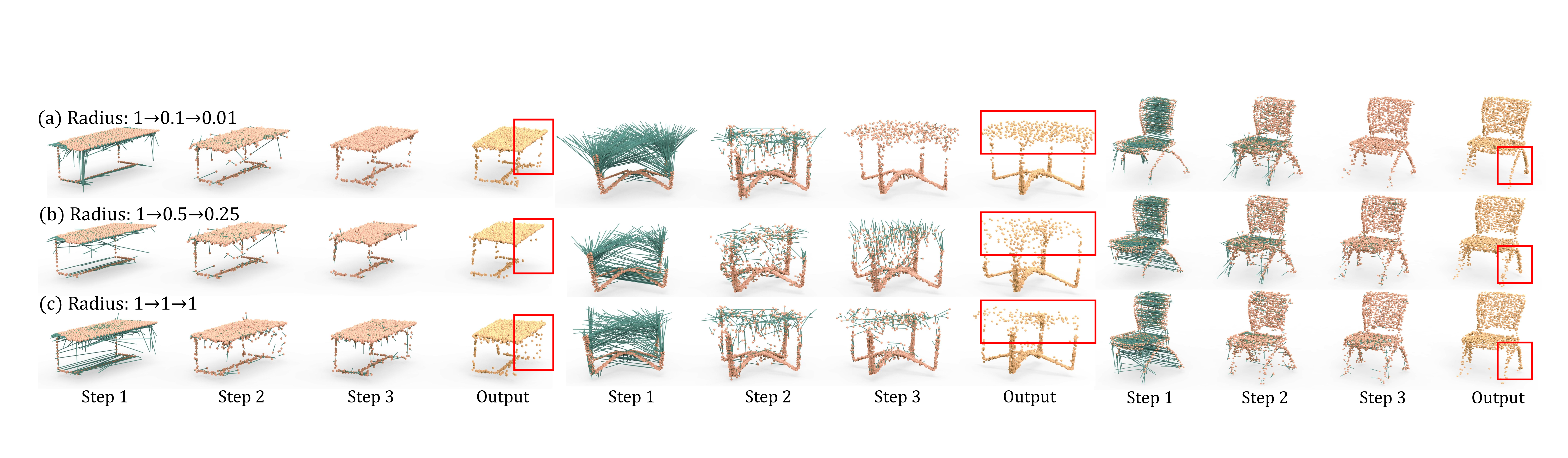}\vspace{-0.4cm}
  \caption{Illustration of deformation process in each step under different strategies of searching radius.
  }
  \label{fig:radius_analysis}
\end{figure*}

\subsubsection{Analysis of searching radius}
By default, we decrease the searching radius for each step by the ratio of 10. In Table \ref{table:radius_analysis}, we analyze different sittings of searching radius and evaluate their influence to the performance of PMP-Net++ framework. We additionally test two different strategies to perform point moving path searching, i.e. the strategy without decreasing searching radius ($[1.0,1.0,1.0]$ for each step), and the strategy with smaller decreasing ratio ($[1.0,0.5,0.25]$). The baseline result is the default setting of PMP-Net++ ($[1.0,0.1,0.01]$ for each step).  Table \ref{table:radius_analysis} shows that both PMP-Net and PMP-Net++ achieve the worst performance at $[1.0,1.0,1.0]$. This is a non-decrease strategy where the searching radius in each step weighs the same to the network, and the better experimental results achieved by the other two variations justify the effectiveness of the strategy to decrease searching radius. And when comparing strategy of $[1.0,0.5,0.25]$ with $[1.0,0.1,0.01]$, we can find that decreasing searching radius with larger ratio can improve the model performance, because larger ratio can better prevent the network from overturning the decisions in previous steps. We also note that when the decreasing ratio becomes large, the PMP-Net++ will approximate the behavior of network with step=1 in Table \ref{table:step_analysis}, which can in return be harmful to the performance of shape completion.
\begin{table}[!h]
\centering
\caption{The effect of searching radius (baseline marked by ``*'').}
\resizebox{\columnwidth}{!}{\begin{tabular}{l|lccccc}
\toprule
Backbone&Radius. &Avg.    &Plane   &Chair  &Car  &Table  \\ \midrule
\multirow{3}{*}{PMP-Net}&$[1.0,1.0,1.0]$  &12.01  &3.61  &16.44 &8.22 &19.79\\
&$[1.0,0.5,0.25]$  &11.77  &\textbf{3.36}   &15.92 &8.01  &19.79\\
&$[1.0,0.1,0.01]$*  &\textbf{11.58}  &3.42   &\textbf{15.88} &\textbf{7.87} &\textbf{19.15}\\\midrule
\multirow{3}{*}{PMP-Net}&$[1.0,1.0,1.0]$  &8.69  &2.58  &12.38 &6.86 &12.90\\
&$[1.0,0.5,0.25]$  &8.54  &\textbf{2.33}  &11.33  &6.66 &13.79\\
&$[1.0,0.1,0.01]$*  &\textbf{8.03}  &2.58  &\textbf{10.70}  &\textbf{6.44} &\textbf{12.30}\\
\bottomrule
\end{tabular}}
\label{table:radius_analysis}
\end{table}

\subsubsection{Visual analysis of point moving path under different radius}
In Figure \ref{fig:radius_analysis}, we visualize the searching process under different strategies of searching radius in Table \ref{table:step_analysis}. By analyzing the deformation output in step 1, we can find that PMP-Net++ with a coarse-to-fine searching strategy can learn to predict a better shape at early step, where the output of step 1 in Figure \ref{fig:radius_analysis} (a) is more complete and tidy than the ones in Figure \ref{fig:radius_analysis} (b) and Figure \ref{fig:radius_analysis} (c). Moreover, a better overall shape predicted in the early stage will enable the network focus on refining a better detailed structure of point cloud, which can be concluded from the comparison of step 3 in Figure \ref{fig:radius_analysis}, where the region highlighted by red rectangles in Figure \ref{fig:radius_analysis} (a) is much better than the other two subfigures.

\subsubsection{Dimension of noise vector}
The noise vector in Eq.(1) in our paper is used to push the points to leave their original place. In this section, we analyze the dimension and the standard deviation of the noise, which may potentially decide the influence of the noise to the points. Because either the dimension or the standard deviation of the noise vector decreases to 0, there will be no disturbance in the network. On the other hand, larger vector dimension or standard deviation will cause larger disturbance in the network. In Table \ref{table:analysis_noiseDim}, we first analyze the influence of dimension of noise vector. By comparing 0-dimension result with others, we can draw conclusion that the disturbance caused by noise vector is important to learn the point deformation. And by analyzing the performance of different length of noise vector, we can find that the influence of vector length is relatively small, compared with the existence of noise vector.

\begin{table}[!h]
\centering
\caption{The effect of noise dimension (baseline marked by ``*'').}
\begin{tabular}{l|lccccc}
\toprule
Backbone &Dim. &Avg.    &Plane   &Car  &Chair  &Table  \\ \midrule
\multirow{5}{*}{PMP-Net}&0  &14.56  &4.39  &10.48  &19.01 &24.33\\
&8  &11.85  &3.28  &7.95  &15.65 &20.50\\
&16  &11.68  &3.44  &\textbf{7.86}  &\textbf{15.22} &20.19\\
&32*  &\textbf{11.58}  &3.42  &7.87  &15.88 &\textbf{19.15}\\
&64  &\textbf{11.58}  &\textbf{3.14}  &7.96  &16.01 &19.17\\\midrule
\multirow{5}{*}{PMP-Net++}&0  &11.10  &3.33  &8.66  &14.63 &17.68\\
&8  &8.20  &2.23  &6.53  &10.80 &13.10\\
&16  &8.24  &2.46  &6.66  &10.72 &13.13\\
&32*  &\textbf{8.03}  &2.58    &\textbf{6.44}  &\textbf{10.70} &\textbf{12.30}\\
&64  &8.11  &\textbf{2.39}  &6.50  &10.80 &12.69\\
\bottomrule
\end{tabular}
\label{table:analysis_noiseDim}
\end{table}

\subsubsection{Standard deviation of noise distribution} In Table \ref{table:analysis_noiseStd}, we show the completion results of PMP-Net++ under different standard deviations of noise vector. Similar to the analysis of vector dimension, we can draw conclusion that larger disturbance caused by bigger standard deviation will help the network achieve better completion performance. The influence of noise vector becomes weak when the standard deviation reaches certain threshold (around $10^{-1}$ according to Table \ref{table:analysis_noiseStd}).
\begin{table}[!h]
\centering
\caption{The effect of standard deviation (baseline marked by ``*'').}
\begin{tabular}{l|lccccc}
\toprule
Backbone &Stddev.     &Avg.    &Plane   &Car  &Chair  &Table  \\ \midrule
\multirow{5}{*}{PMP-Net}&$10^{-2}$   &11.89  &3.32  &8.15  &16.42 &19.58\\
&$10^{-1}$   &\textbf{11.56}  &3.58  &7.78  &15.47 &19.41\\
&$1.0$*       &11.58  &3.42  &\textbf{7.87}  &15.88 &\textbf{19.15}\\
&$10$        &11.62  &\textbf{3.35}  &7.88  &\textbf{15.29} &19.95\\\midrule
\multirow{5}{*}{PMP-Net++}&$10^{-2}$   &8.08  &2.74  &6.54  &\textbf{10.60} &12.31\\
&$10^{-1}$   &8.43  &2.35  &6.56  &10.81 &13.89\\
&$1.0$*       &\textbf{8.03}  &2.58    &\textbf{6.44}  &10.70 &\textbf{12.30}\\
&$10$        &8.25  &\textbf{2.27}  &6.49  &11.30 &12.82\\
\bottomrule
\end{tabular}
\label{table:analysis_noiseStd}
\end{table}

\subsubsection{Efficiency analysis}
The efficiency analysis on Completion3D dataset is given in Table \ref{table:efficiency_analysis}
We can find that PMP-Net++ is more efficient than GRNet in terms of both parameters and FLOPs, which proves our opinion that voxel-aided methods may suffer from the computational inefficiency problem. Moreover, the efficiency of PMP-Net++ is also comparable with PCN. Since PCN adopts a linear layer to generate coarse point cloud in its network designation, it still requires lots of computational resources even in single-step point cloud completion
\begin{table}[!h]
\centering
\caption{Comparison with deformation from grid points.}
\begin{tabular}{l|ccc}
\toprule
Methods &PCN   &GRNet   &PMP-Net++(3 steps)  \\ \midrule
Params(M) &5.29  &76.7  &5.89 \\
FLOPs(G) &3.31  &18.6  &4.48 \\
\bottomrule
\end{tabular}
\label{table:efficiency_analysis}
\end{table}

\section{Conclusions}
In this paper, we propose a novel PMP-Net++ for point cloud completion by multi-step shape deformation. By moving points from the source to the target point clouds with multiple steps, PMP-Net++ can consistently refine the detailed structure and topology of the predicted shape, and establish the point-level shape correspondence between the incomplete and the complete shape. In experiments, we show the superiority of PMP-Net++ by comparing with other methods on the Completion3D benchmark and PCN dataset, and also demonstrate its good performance in the point cloud up-sampling task.

In all, the research of PMP-Net++ is somewhat limited by the lack of effective constraints to the deformation process. In this paper, although the PMD loss has been proposed to regularize the moving distances of all points in 3D space, the results may not be satisfactory in some examples due to the lack of supervision like point moving directions or final destinations. Therefore, the network may have difficulty to learn the optimal solution during the training process. In our opinion, a potential solution to this problem is to further explore the PMD loss, which we plan to try in our future work. This solution not only constrains the total moving distance, but also can guide the network to learn the directions of each move.
\bibliographystyle{IEEEtran}
\bibliography{ref}

\begin{IEEEbiography}[{\includegraphics[width=1in,height=1.25in,clip,keepaspectratio]{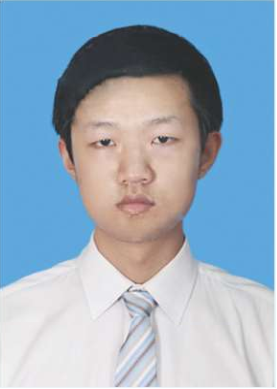}}]{Xin Wen}
received the B.S. degree in engineering management from the Tsinghua University, China, in 2012. He is currently the PhD student with the School of Software, Tsinghua University. His research interests include deep learning, shape analysis and pattern recognition, and NLP.
\end{IEEEbiography}

\begin{IEEEbiography}[{\includegraphics[width=1in,height=1.25in,clip,keepaspectratio]{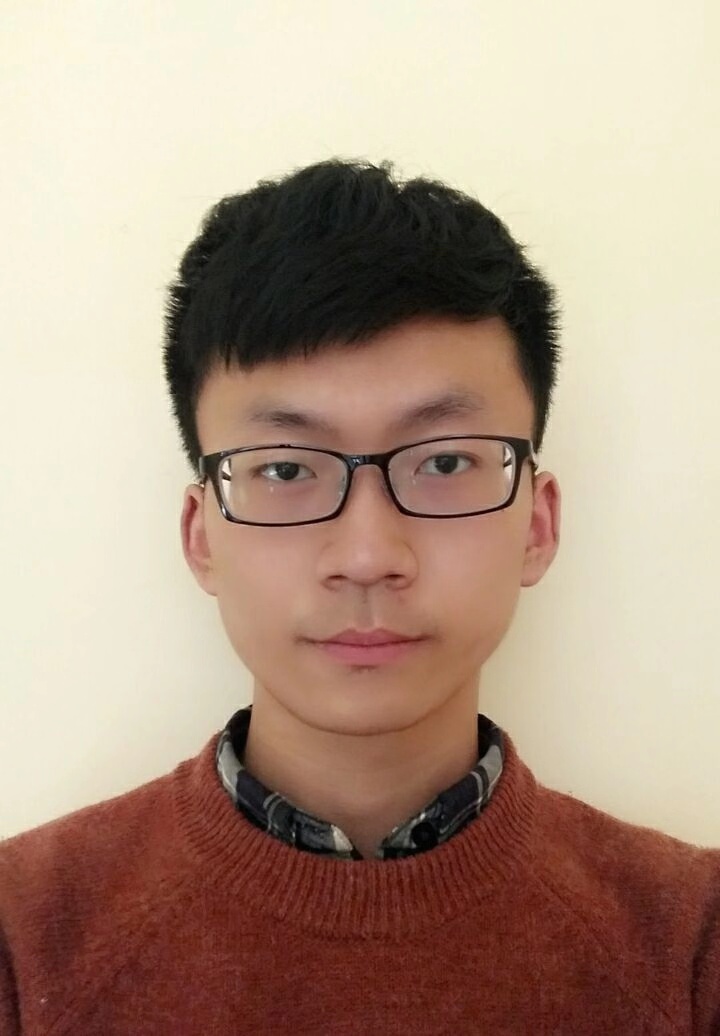}}]{Peng Xiang}
received the B.S. degree in software engineering from Chongqing University, China, in 2019. He is currently the graduate student with the School of Software, Tsinghua University. His research interests include deep learning, 3D shape analysis and pattern recognition.
\end{IEEEbiography}

\begin{IEEEbiography}[{\includegraphics[width=1in,height=1.25in,clip,keepaspectratio]{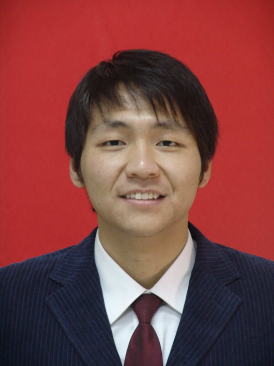}}]{Zhizhong Han}
received the Ph.D. degree from Northwestern Polytechnical University, China, 2017. He was a Post-Doctoral Researcher with the Department of Computer Science, at the University of Maryland, College Park, USA. Currently, he is an Assistant Professor of Computer Science at Wayne State University, USA. His research interests include 3D computer vision, digital geometry processing and artificial intelligence.
\end{IEEEbiography}

\begin{IEEEbiography}[{\includegraphics[width=1in,height=1.25in,clip,keepaspectratio]{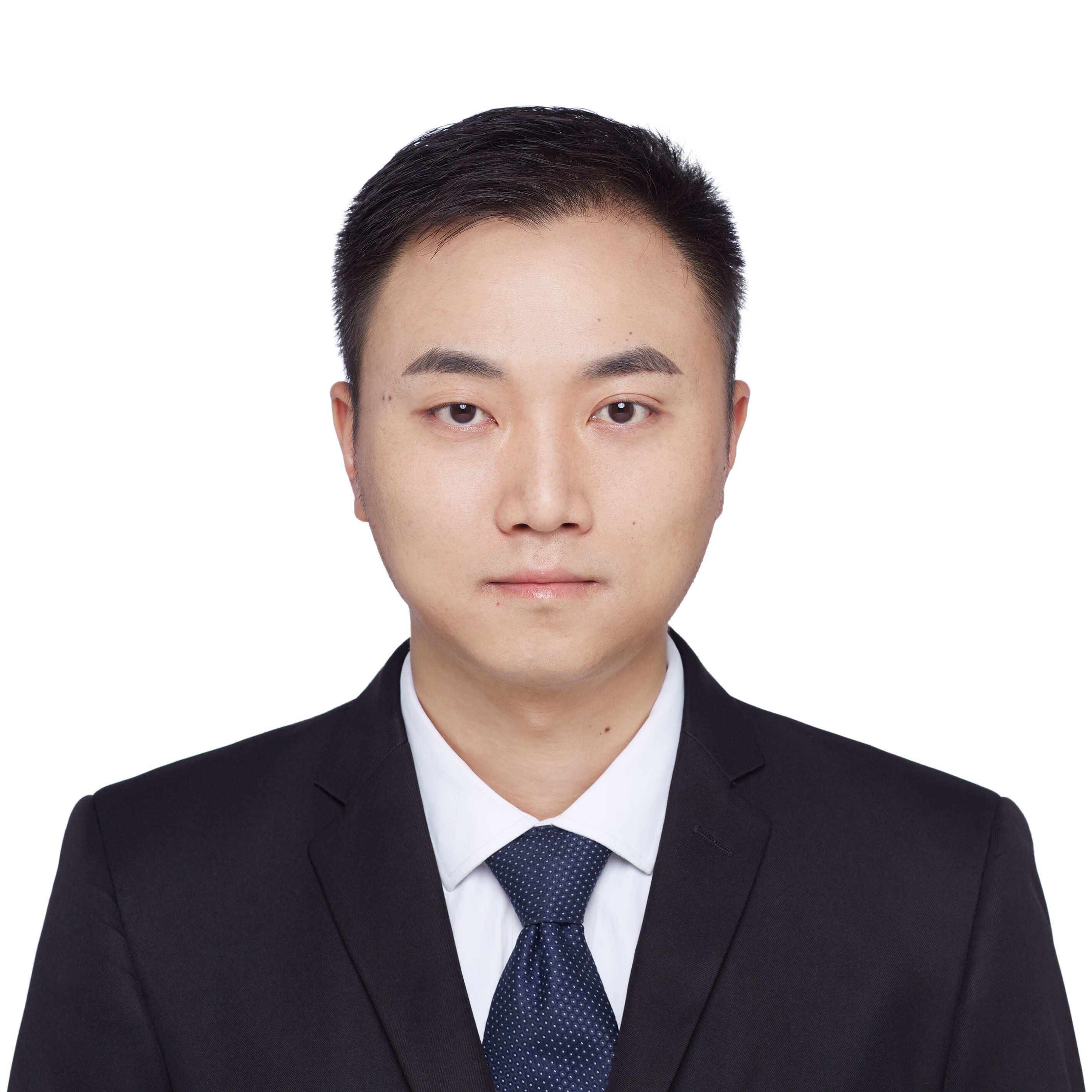}}]{Yan-Pei Cao}
received the bachelor's and Ph.D. degrees in computer science from Tsinghua University in 2013 and 2018, respectively. He is currently a research engineer at Y-tech, Kuaishou Technology. His research interests include geometric modeling and processing, 3D reconstruction, and 3D computer vision.
\end{IEEEbiography}

\begin{IEEEbiography}[{\includegraphics[width=1in,height=1.25in,clip,keepaspectratio]{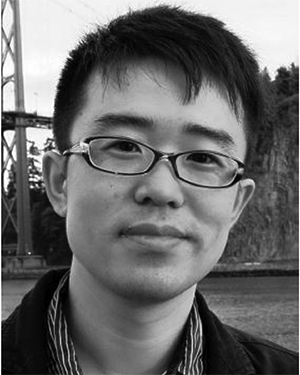}}]{Pengfei Wan}
received the B.E. degree in electronic engineering and information science from the University of Science and Technology of China, Hefei, China, and the Ph.D. degree in electronic and computer engineering from the Hong Kong University of Science and Technology, Hong Kong. His research interests include image/video signal processing, computational photography, and computer vision.
\end{IEEEbiography}

\begin{IEEEbiography}[{\includegraphics[width=1in,height=1.25in,clip,keepaspectratio]{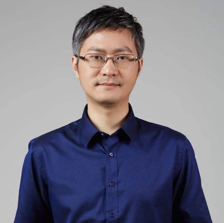}}]{Wen Zheng}
received the bachelor's and master's degrees from Tsinghua University, Beijing, China, and the Ph.D. degree in computer science from Stanford University, in 2014. He is currently the Head of Y-tech at Kuaishou Technology. His research interests include computer vision, augmented reality, machine learning, and computer graphics.
\end{IEEEbiography}

\begin{IEEEbiography}[{\includegraphics[width=1in,height=1.25in,clip,keepaspectratio]{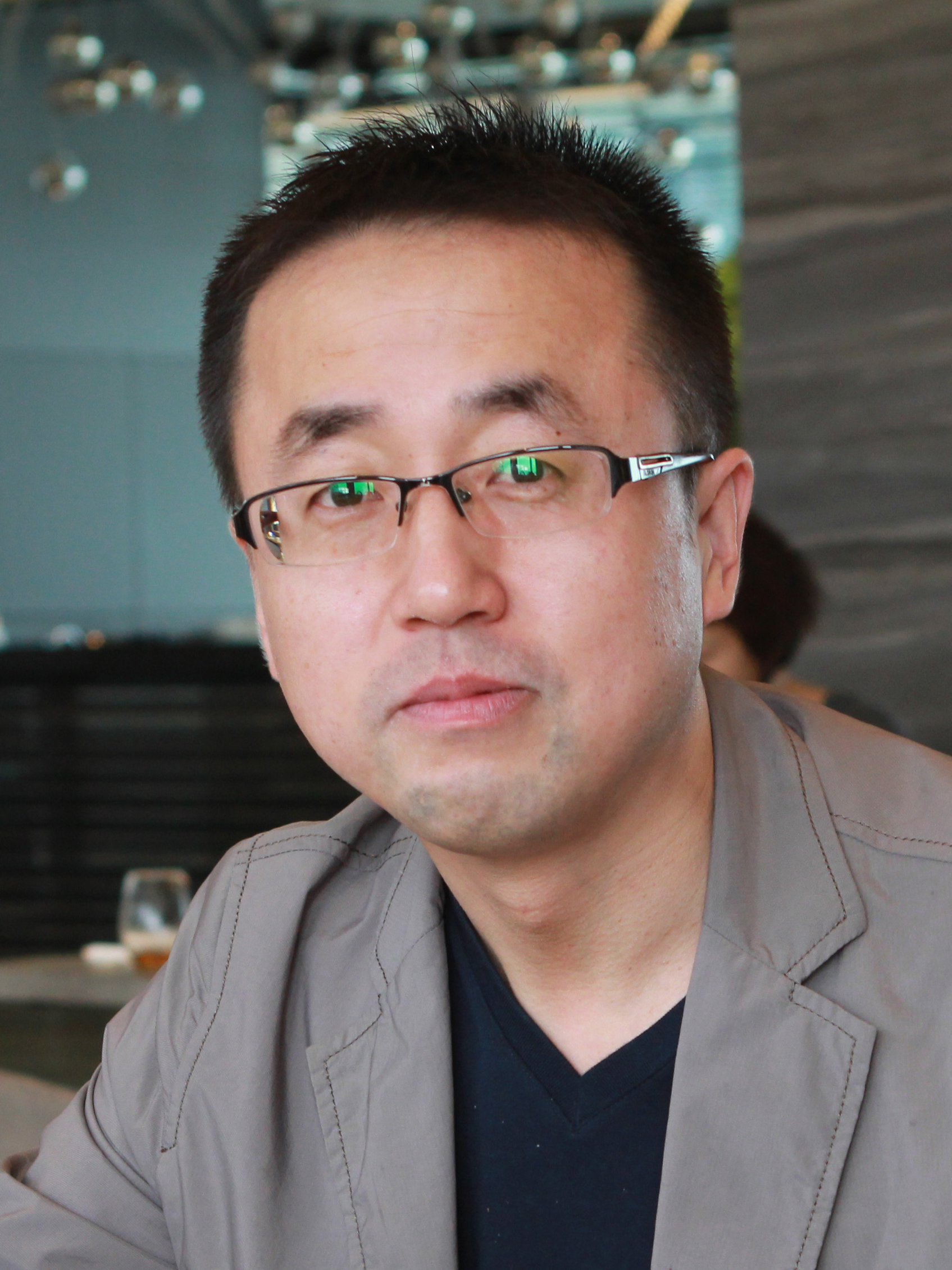}}]{Yu-Shen Liu}
(M'18) received the B.S. degree in mathematics from Jilin University, China, in 2000, and the Ph.D. degree from the Department of Computer Science and Technology, Tsinghua University, Beijing, China, in 2006. From 2006 to 2009, he was a Post-Doctoral Researcher with Purdue University. He is currently an Associate Professor with the School of Software, Tsinghua University. His research interests include shape analysis, pattern recognition, machine learning, and semantic search.
\end{IEEEbiography}




\end{document}